\documentclass[fleqn,a4paper,10pt]{article}

\usepackage{amsmath,amssymb,amsfonts}
\usepackage{mathtools}
\usepackage{amsthm}
\usepackage{physics}
\usepackage{bm}
\usepackage{tikz}
\usepackage{empheq}
\usepackage{mdframed}
\usepackage{multirow}
\usepackage{kbordermatrix}
\usepackage{relsize}

\newcommand{\dist}{{\textstyle \mathsmaller{\varDelta}}} 

\let\originalleft\left
\let\originalright\right
\renewcommand{\left}{\mathopen{}\mathclose\bgroup\originalleft}
\renewcommand{\right}{\aftergroup\egroup\originalright}

\theoremstyle{definition}

\newmdtheoremenv{res}{Results}

\title{Covariance and Correlation Measures on a Graph in a Generalized Bag-of-Paths Formalism
\quad \\
  \normalsize{\textit{Draft manuscript subject to changes}}}

\author{Guillaume Guex \\
		{\small SLI, Universit\'{e} de Lausanne, Switzerland} \and Sylvain Courtain \\
		{\small LOURIM, Universit\'{e} catholique de Louvain, Belgium} \and Marco Saerens \\	{\small ICTEAM, Universit\'{e} catholique de Louvain, Belgium\footnote{Marco Saerens is also fellow researcher at the Universit\'{e} Libre de Bruxelles (IRIDIA Laboratory), Belgium.}}}

\begin{document}

\date{}
\maketitle

\begin{abstract}
{This work derives closed-form expressions computing the expectation of co-presence and of number of co-occurrences of nodes on paths sampled from a network according to general path weights (a bag of paths). The underlying idea is that two nodes are considered as similar when they often appear together on (preferably short) paths of the network. The different expressions are obtained for both regular and hitting paths and serve as a basis for computing new covariance and correlation measures between nodes, which are valid positive semi-definite kernels on a graph. Experiments on semi-supervised classification problems show that the introduced similarity measures provide competitive results compared to other state-of-the-art distance and similarity measures between nodes.}
{Link analysis; network data analysis; complex networks; network science; graph mining; kernel on a graph; bag-of-paths model.}
\end{abstract}


\section{Introduction}


\subsection{General introduction}

This work addresses the important problem of defining similarities and distances between nodes of a network based on its structure, faced in many applications such as link prediction, community detection, node classification, and network visualization, among others \cite{barabasi2016network,chiang2012networked,chung2006complex,estrada2012structure,Fouss-2016,kolaczyk2009statistical,lewis2009network,mihalcea2011graph,Newman-2018,silva2016machine,thelwall2004link,Wasserman-1994}. It extends previous work on the \emph{randomized shortest paths} and the \emph{bag-of-paths} frameworks introduced in a series of previous papers \cite{bavaud2012interpolating,Francoisse-2017,Kivimaki-2012,Saerens-2008,Yen-08K}, and most notably \cite{Mantrach-2009}. This effort was initially inspired by models developed in transportation science, especially \cite{Akamatsu-1996,Dial71}.

 Of course, different meaningful notions of similarity between nodes can be defined, depending on the application \cite{Wasserman-1994}. The most common one states that two nodes are considered as similar if (\textit{i}) they are both close in the network (in terms of shortest path distance) and highly inter-connected. In other words, two nodes are similar when they are highly accessible from each other \cite{Chebotarev-1997,Chebotarev-1998a,Lebichot-2018}. Another approach (\textit{ii}) considers that two nodes are similar when they share some common properties, for instance they often co-occur on paths or trees sampled from the network. This is the approach that will be investigated in this paper by considering a general bag-of-paths approach. A third relevant idea states that (\textit{iii}) two nodes could be considered as related when they play comparable roles in the network, for instance if they influence the network in a similar way.

In this context, as discussed in \cite{Lebichot-2018} (see also \cite{Fouss-2016,Francoisse-2017}), the specificity of our approach can be understood as follows.
Most traditional distances or similarity measures between nodes are based on two common paradigms about the transfer of information, or more generally the movement, occurring in the network: optimal communication based on shortest paths and random communication based on a random walk on the graph. For instance, the shortest path distance is based on geodesics and the resistance distance (\cite{klein1993resistance}, proportional to the commute-time distance \cite{Chandra-1989,FoussKDE-2005}), on random walks. However, both the shortest path and the resistance distance suffer from some annoying drawbacks \cite{Fouss-2016}: the shortest path distance does not integrate the amount of connectivity between the two nodes and produces many ties in unweighted networks, whereas random walks quickly loose the notion of proximity to the initial node when the graph becomes larger \cite{vonLuxburg-2010,Luxburg-2014}.

Contrarily to these standard measures, the randomized shortest paths framework integrates both proximity and amount of connectivity for defining distance and betweenness measures between pairs of nodes based on approach (\textit{i}) \cite{yen2008family,Saerens-2008,Kivimaki-2012,Lebichot-2014,Fouss-2016}. In its basic form, this model assumes that paths connecting the two nodes are chosen according to a Gibbs-Boltzmann probability distribution depending on a temperature parameter balancing the smoothness of the measure. When the temperature of the model is high, communication occurs through a random walk while, for low temperatures (close to zero), shorter paths are promoted. The model has been extended recently by adding a priori probabilities on the starting and ending nodes, thus allowing to weigh nodes \cite{guex2016interpolating,Guex-2019}.

In the same spirit, the bag-of-paths based measures aim to quantify the similarity between the nodes by considering arbitrary paths between all pairs of nodes (and not only between one pair of predefined nodes) favoring low-cost, and thus short, paths \cite{Mantrach-2009,Francoisse-2017}. Similarity measures can then be derived by computing number of co-occurrences of nodes on paths, as in approach (\textit{ii}). The present work expands this approach by considering a more generic framework and deriving new measures based on node presence and on hitting paths in a systematic way. A nice property is that all the quantities of interest can be computed in closed form by standard matrix operations. Moreover, the introduced measures have nice, intuitive, interpretations, as explained in the next subsection.

Other such families of distances were recently introduced and studied, for instance in \cite{Chebotarev-2011,Chebotarev-2012,Chebotarev-2013} by considering the co-occurrences of nodes in forests or walks on a graph, in \cite{alamgir2011phase,Herbster-2009,Luxburg-2014} based on a generalization of the effective resistance in electric circuits, in \cite{Li-2011,Li-2013} by flow optimization with mixed L1-L2 norms, and in \cite{bavaud2012interpolating,guex2015flow} by considering network flows with entropy regularization.
The originality of our work, in comparison with these previously developed methods, lies in the fact that we adopt a \emph{bag-of-paths} formalism; that is, the quantities of interest are defined on the set of whole paths (or walks, trajectories) appearing in the network.

For a comprehensive survey of related work on the design of similarity/distance measures on graphs and networks, see \cite{FoussKernelNN-2011,Francoisse-2017,Kivimaki-2012,Mantrach-2009} as well as \cite{Fouss-2016}. However, three closely related and highly relevant works must be emphasized here. The first work \cite{Chebotarev-2002b,Chebotarev-1997,Chebotarev-1998a} develops new similarity measures based on the co-occurrence of nodes on the same tree appearing on forests (sets of trees) sampled from the graph. The main quantity is called the relative forest accessibility between nodes and measures to which extend two nodes co-occur on the same tree, and are therefore both close and highly connected. The approach is similar to (and has inspired) our work but involves different motifs: trees instead of paths.
Another interesting closely related work is \cite{Kolaczyk-2009c} which introduces the concept of co-betweenness. The authors aim to extend node betweenness centrality to sets of nodes in terms of shortest paths that pass through all nodes in a set. They then provide an expansion for group betweenness in terms of increasingly higher orders of co-betweenness. Moreover, they derive an efficient algorithm for computing the pairwise co-betweenness involving two nodes only, which generalizes the standard shortest-path node betweenness. Our approach is similar in spirit, but is based on paths sampled from the graph according to a general probability distribution favoring low-cost paths, instead of shortest paths.
Finally, the recent DeepWalk \cite{Perozzi-2014} is based on simulating a large number of walks in the graph and use the vord2vec approach to compute a $p$-dimensional representation of the nodes based on their co-occurrences in a sliding time frame. The algorithm provides a graph representation (or embedding) capturing the association between the nodes in terms of co-occurrence. In the present paper, the same idea is exploited, but the association measures between nodes (correlations and covariances) are computed in closed form instead of relying on computer-based path generation. The DeepWalk approach has become very popular in the deep learning community.


\subsection{The main intuition behind the models}
\label{Subsec_intuition_models01}

The idea behind the introduced node similarity measures is as follows. Let us assume we can enumerate all paths $\wp \in \mathcal{P}$ (the set of all paths) that can be sampled from a network $G$. Further consider that a measure of ``quality" of these paths can be easily computed, the weight $w(\wp)$, as, e.g., in \cite{Chebotarev-2012}. This weight reflects the reward of following the path, and could be based on, e.g., the length of the path, the total cost along the path, the time to follow the path, etc.
Moreover, let us assume that paths are indexed and ordered by decreasing degree of quality, $(\wp_{1},\wp_{2},\dots)$. As an example let us consider a corpus of documents from which a network is extracted (see e.g., \cite{mihalcea2011graph}). If the nodes of the network represent terms and the (weighted) links represent co-occurrences of terms within a given window in the text, then paths in this network correspond to sequences of words (sentences) where highly likely sequences have a higher weight.

The paths selection strategy, based on a sampling probability distribution derived from path weights, naturally favours high-weight paths and the overall quality level of the chosen paths could be monitored thanks to a temperature parameter (see Subsection \ref{classical_BoP}), depending on the problem. Then, if the parameter is close to zero, only high-quality paths are considered while, for high values, all paths are more and more considered equally. The model can thus be considered as a bag of paths from which paths are drawn \cite{Mantrach-2009,Francoisse-2017}.
Within this context, and as already mentioned, our similarity measure has the following interpretation: two nodes are considered as highly similar when they often co-occur on the same paths, when drawn thanks to a probability distribution favouring high-weight paths. In other words, two nodes are considered as related if they share the same paths.

Intuitively, this can be captured in the following way based on the enumeration of paths, although some tricks will be used in order to efficiently compute the quantities. Let $\mathbf{X}$ be a data matrix (the path-node matrix, inspired by the document-term matrix in information retrieval) with rows corresponding to paths, $\wp_{i} \in \mathcal{P}$, and columns to nodes, $j \in \{ 1 \cdots n \}$, of $G$. The entries $i,j$ of this matrix are binary with a $1$ if the node $j$ appears on the path $\wp_{i}$ and zero otherwise.

Moreover, we further introduce a diagonal matrix weighting the paths, and containing as diagonal elements $i,i$ the probability $\mathrm{P}(\wp_{i})$ of choosing path $\wp_{i}$ according to the sampling distribution with $\sum_{i=1}^{\infty} \mathrm{P}(\wp_{i}) = 1$. Thus, as an example, these two matrices are of the form:
\begin{align*}
&\mathbf{X} =
\kbordermatrix{
  &  1  &  2  &  3  & \cdots  \cr
\wp_{1} &  1  &  0  &  1  & \cdots  \cr
\wp_{2} &  0  &  1  &  0  & \cdots  \cr
\wp_{3} &  1  &  0  &  1  & \cdots  \cr
\wp_{4} &  0  &  0  &  0  & \cdots  \cr
\vdots  & \vdots  & \vdots  & \vdots & \ddots  \cr
}; \quad
\mathbf{D} =
\kbordermatrix{
  &  \wp_{1}  &  \wp_{2}  &  \wp_{3}  &  \wp_{4}  & \cdots  \cr
\wp_{1} &  \mathrm{P}(\wp_{1})  &  0  &  0  &  0  & \cdots  \cr
\wp_{2} &  0  &  \mathrm{P}(\wp_{2})  &  0  &  0  & \cdots  \cr
\wp_{3} &  0  &  0  &  \mathrm{P}(\wp_{3})  &  0  & \cdots  \cr
\wp_{4} &  0  &  0  &  0  &  \mathrm{P}(\wp_{4})  & \cdots  \cr
\vdots  & \vdots  & \vdots  & \vdots  & \vdots & \ddots  \cr
}
\end{align*}
In this example, we observe that node $1$ and node $3$ look similar as they appear on the same paths ($\wp_{1}$ and $\wp_{3}$). Conversely, nodes $1$ and $2$ appear quite different.

In other words, nodes are characterized by their appearance on paths so that a binary feature vector indicating their presence on the different paths, $\mathbf{x}_{j}$, is associated to each node\footnote{Note that all vectors are considered as column vectors.}. The $\mathbf{x}_{j}$ vectors can therefore be viewed as profile vectors characterizing each node $j$ with respect to its presence on paths. Thus, these vectors form the column vectors of the data matrix $\mathbf{X}$.
Alternatively, the data matrix could also contain the number of occurrences of each node on the different paths, instead of a binary presence value.

Now, a simple, but still meaningful, measure of the similarity between pairs of nodes $i$ and $j$ is simply the expected frequency of common presence on the paths. This quantity can be computed by taking the inner product $\mathbf{x}_{i}^{\mathrm{T}} \mathbf{D} \mathbf{x}_{i}$ for node $i$, or $\mathbf{X}^{\mathrm{T}} \mathbf{D} \mathbf{X}$ for the result on all pairs of nodes.
Such quantities will be studied and computed in this paper in a relatively general setting.

Note that this kind of similarity measure is closely related to contextual similarities used in, e.g., information retrieval where two words are considered as related when they often appear in the same context (same sentence, window or document) \cite{Harispe-2015}, as illustrated by the recent, popular, vord2vec method \cite{Mikolov-2013} and the emerging field of representation learning \cite{Bengio-2013,Zhang-2018}. Similarly, the context is represented here by paths of arbitrary length and the measures of association are computed directly in closed form from the structure of the graph and its edge weights.
As an example, and as already mentioned, this idea has been exploited recently in the field of deep learning for computing a graph representation \cite{Perozzi-2014} and is now very popular in that field.


\subsection{Contributions and contents}

The paper derives closed-form expressions for computing similarity measures for two types of paths: (\textit{i}) \emph{regular} (non-hitting) and (\textit{ii}) \emph{hitting} paths. More precisely, the derived similarities are the \emph{covariance} and \emph{correlation} kernels on a graph (they are positive semidefinite). Moreover, these kernels are defined for two different measures: based on (\textit{i}) simple binary common \emph{presence} of nodes on paths and (\textit{ii}) \emph{number of co-occurrences} on paths.
In addition, various betweenness centrality measures are also derived within the framework.

 
The material presented in this paper is therefore an extension of the previous work \cite{Mantrach-2009} where a covariance and a correlation kernel were derived for regular paths based on number of co-occurrences only. The derivation of the different quantities in the present work is more systematic, comprehensive, and generic. Moreover, the results involving node presence are new and use a completely different technique than for those based on number of co-occurrences. The same is true for the results involving hitting paths (instead of regular paths), which are more elaborate. Finally, the framework developed in this work is more general as it can be applied to a large class of weights defined on edges, whereas it was restricted to the bag-of-paths model based on Kullback-Leibler divergence regularization in \cite{Mantrach-2009}. The introduced framework contains the standard bag-of-paths model as a special case.

The paper is organized as follows. First, the underlying background, notation and framework are detailed in Section \ref{Sec_main_framework}. Then, Section \ref{section_weight_formulae} derives various important expressions computing the weights on sets of paths avoiding or containing some nodes. These results are derived for both regular and hitting paths and provide the basic support for the definition of the similarity measures. Section \ref{copre_cooc_formulae} develops betweenness centrality and node similarity measures based on the presence and the number of occurrences of nodes on paths, for both regular and hitting paths. Finally, those measures are assessed and compared in Section \ref{case_studies}. Section \ref{Sec_conclusion} is the conclusion.


\section{Framework and notation}
\label{Sec_main_framework}


\subsection{The generalized bag-of-paths formalism}
\label{Subsec_generalized_BoP_framework}

As already stated in the introduction, the standard \emph{bag-of-paths} framework (BoP)  \cite{Francoisse-2017,Mantrach-2009} sets up a \emph{Gibbs-Boltzmann} distribution defining the probabilities of drawing a path from the \emph{set of all paths in the graph}, also named the \emph{bag of paths}, by assigning higher probabilities to short paths and lower probabilities to long paths. The standard bag-of-baths formalism will be described in more details in Section \ref{classical_BoP}. For the moment, we will define a more generic framework, namely the \emph{generalized bag-of-paths formalism}, inspired by \cite{Chebotarev-2012}. A summary of main notations appears in Table \ref{Tab_notation01}.


\subsubsection{Paths, hitting-paths, and sets of paths}

Let $G = (\mathcal{V},\mathcal{E})$ be a weighted, strongly connected, directed graph, with set of nodes $\mathcal{V} = \{1,2,\ldots,n\}$ and set of edges $\mathcal{E} = \{(i,j)\}$ containing $m$ elements. This graph is described by its \emph{weighted adjacency matrix} $\mathbf{W} =(w_{ij})$ (or simply \emph{weight matrix}), representing non-negative local affinities between nodes or rewards on edges, with $w_{ij} \geq 0$. Moreover, the weights must be equal to zero for missing links: $w_{ij} = 0$ when there is no link between $i$ and $j$. This weight matrix is not simply the usual adjacency matrix $\mathbf{A}$ in general, but a function of it depending on the considered application like, e.g., a substochastic transition matrix representing a killed random walk (see the end of Subsection \ref{Subsec_generalized_BoP_framework} for some examples). The weight matrix will be used in order to define \emph{path weights} and \emph{path probabilities} on sets of paths. We will see later that it must satisfy some simple constraint in order to make sure that probabilities of drawing paths are well-defined: it should have a \emph{spectral radius} strictly lower than $1$. This point is discussed in Section \ref{W_radius}.

A $\ell$-length (regular) \emph{path} on the graph $G$, denoted by $\wp$, is a sequence of nodes $\wp = (i_0,i_1,\dots,i_{\ell-1},i_{\ell})$, where $\ell \geq 0$ and $(i_\tau,i_{\tau+1}) \in \mathcal{E}$ for all $\tau = 0,\dots,\ell-1$. Note that $0$-length paths are allowed by convention. We denote by the variable $\wp_{st}$ a path whose starting node is $s$ and ending node is $t$. A \emph{hitting path}, denoted by the superscript $\mathrm{h}$ in $\wp^\mathrm{h}$, is defined as a path such that the final node $i_{\ell}$ appears only once on the path, i.e., $i_{\ell} \neq i_{\tau},\, \forall \tau = 1,\dots,\ell-1$ and of $0$ length if $i_0 = i_{\ell}$. In other words, for hitting paths, the final node is considered as an absorbing node: it can thus only appear once, at the end of the path.

The \emph{set of all paths} in $G$, also called the \emph{bag-of-paths}, is denoted by $\mathcal{P}$. This set $\mathcal{P} = \cup_{s,t=1}^{n} \mathcal{P}_{st}$ is the union of the subsets $\mathcal{P}_{st}$ of all regular (non-hitting) paths starting in node $s$ and ending in target node $t$. Several subsets of the bag of paths will be used in the sequel and, for the sake of clarity in further developments, we will use special symbols for these different subsets.

The superscript in $\mathcal{P}^\mathrm{h}$ will refer to the set of \emph{hitting paths}, also named the \emph{bag-of-hitting-paths}. Still another type of superscript has the form $\mathcal{P}^{(+\mathcal{I})}$ or $\mathcal{P}^{(-\mathcal{I})}$, where $\mathcal{I} \subset \mathcal{V}$ is a subset of nodes. This superscript indicates that the set is composed of paths \emph{containing} (with a $+$ symbol), and respectively \emph{avoiding} (with a $-$), all of the nodes in $\mathcal{I}$. Note that we will usually simply write $\mathcal{P}^{(+i)}$ instead of $\mathcal{P}^{(+\{i\})}$ when there is only one node $i$ in the set. Of course, all these notations can be combined. For example, $\mathcal{P}^{\mathrm{h}(-\{i,j\})}_{st}$ refers to the subset of hitting paths connecting $s$ to $t$ and avoiding nodes $i$ and $j$.

Note that the set of all paths $\mathcal{P}$ is equipped with a composition rule for two paths where the ending node of one path corresponds to the starting node of the other, i.e., if $\wp_{sk} =  (s,i_1, \ldots, k )$ and $\wp_{kt} = (k,j_1, \ldots, t )$, then $\wp_{sk} \circ \wp_{kt} = (s,i_1, \ldots, k, j_1, \ldots, t )$. By extension, let $\mathcal{P}_{sk} \circ \mathcal{P}_{kt}$ denote the set of all compositions between sets of paths $\mathcal{P}_{sk}$ and $\mathcal{P}_{kt}$.


\begin{table}[t]
\begin{center}
\footnotesize
\begin{tabular}{|l|l|}
\hline
$w_{ij} = [\mathbf{W}]_{ij}$ & element $i,j$ of the weight matrix of graph $G$ \\\hline
$\wp$ & a particular path visiting nodes $s = i_0, i_1, \dots, i_{\ell} = t$ \\\hline
$w(\wp) = \prod_{\tau = 0}^{\ell-1} w_{i_{\tau},i_{\tau+1}}$ & the total weight along path $\wp$ (product of edge weights) \\\hline
$w({\mathcal{Q}}) = \sum_{\wp \in \mathcal{Q}} w(\wp)$ &  the total weight for paths $\wp$ in set of paths $\mathcal{Q}$, $\wp \in \mathcal{Q}$  \\\hline
$\mathcal{P}_{st}(\ell)$     & set of regular paths connecting $s$ to $t$ in exactly $\ell$ steps \\\hline 
$\mathcal{P}_{st}$        & set of regular paths of arbitrary length connecting $s$ to $t$ \\\hline
$\mathcal{P}^{(+i)}_{st}$        & set of regular paths from $s$ to $t$ visiting intermediate node $i$ \\\hline 
$\mathcal{P}^{(-i)}_{st}$        & set of regular paths from $s$ to $t$ avoiding node $i$ \\\hline 
$\mathcal{P} = \cup_{s,t = 1}^{n} \mathcal{P}_{st}$ & set of all regular paths of arbitrary length \\\hline
$\mathcal{P}^{(+\mathcal{I})}$        & set of regular paths from $s$ to $t$ visiting all nodes in set $\mathcal{I}$ \\\hline 
$\mathcal{P}^{(-\mathcal{I})}$        & set of regular paths from $s$ to $t$ avoiding all nodes in set $\mathcal{I}$ \\\hline 
$\mathcal{P}^{\mathrm{h}}_{st}$        & set of hitting paths of arbitrary length connecting $s$ to $t$ \\\hline 
$\mathcal{P}^{\mathrm{h}} = \cup_{s,t = 1}^{n} \mathcal{P}^{\mathrm{h}}_{st}$ & set of all hitting paths of arbitrary length \\\hline
$\mathcal{P}^{\mathrm{h}(+i)}_{st}$        & set of hitting paths from $s$ to $t$ visiting intermediate node $i$ \\\hline 
$\mathcal{P}^{\mathrm{h}(-i)}_{st}$        & set of hitting paths from $s$ to $t$ avoiding node $i$ \\\hline
$\mathcal{P}^{\mathrm{h}(+\mathcal{I})}$        & set of hitting paths from $s$ to $t$ visiting all nodes in set $\mathcal{I}$ \\\hline 
$\mathcal{P}^{\mathrm{h}(-\mathcal{I})}$        & set of hitting paths from $s$ to $t$ avoiding all nodes in set $\mathcal{I}$ \\\hline 
\end{tabular}
\caption{Summary of notation for the enumeration of paths in a graph $G$ for both regular and hitting paths.}
\label{Tab_notation01}
\end{center}
\end{table}


\subsubsection{Paths weights}

Paths weights are defined from the $n \times n$ non-negative weighted adjacency matrix $\mathbf{W}$ of the graph (an example is provided later) and are used to define the bag-of-paths probabilities. The \emph{weight} of a path $\wp = (i_0,\dots,i_{\ell})$, noted $w(\wp)$, is defined as the \emph{product of the weights on its edges}, i.e.,
\begin{equation}
w(\wp) \triangleq \prod_{\tau = 0}^{\ell-1} w_{i_{\tau},i_{\tau+1}}.
\label{Eq_def_path_weight01}
\end{equation}
where we recall that $\ell$ is the length (number of edges) of the path.
By convention, we assume that all $0$-length paths have a weight of one.
Note that this definition favors shorter paths over longer ones because it will be shown below that the property $\mathbf{W}^{\tau} \overset{\tau \rightarrow \infty}{\longrightarrow} 0$ of the weight matrix must hold for consistency. Therefore, the weight of a path decreases with its length, at least in the long term.

We also define the weight of any subset of the bag of paths $\mathcal{Q} \subseteq \mathcal{P}$ as the sum of the weights of its elements (paths),
\begin{equation}
w(\mathcal{Q}) \triangleq \sum_{\wp \in \mathcal{Q}} w(\wp).
\end{equation}
Therefore, if two subsets, $\mathcal{Q}$ and $\mathcal{R}$, are disjoint, we have $w(\mathcal{Q} \cup \mathcal{R}) = w(\mathcal{Q}) + w(\mathcal{R})$.


\subsubsection{Bag-of-paths probabilities}
\label{bop_prob}

We now consider the bag of paths $\mathcal{P}$ and we would like to draw paths from $\mathcal{P}$ with probabilities $\text{P}(\wp)$ proportional to path weights. This is easily obtained by normalizing the weight of the path by the total weight of the bag-of-paths. In short, the  \emph{generalized bag-of-paths probabilities} of drawing a path $\wp$ are defined as
\begin{equation}
\text{P}(\wp) \triangleq \frac{w(\wp)}{\sum_{\wp' \in \mathcal{P}} w(\wp')} = \frac{w(\wp)}{w(\mathcal{P})}.
\label{Eq_pathProbability01}
\end{equation}
Note that the generalized bag-of-paths probabilities $\text{P}(\wp)$ are non-null if and only if the normalizing constant, the weight of the bag-of-paths, i.e., $w(\mathcal{P})$, is finite. This kind of measure is the object of interest of the present work. We will detail in the next section what are the requirements on the weights in order to satisfy this property.

More generally, if we want the conditional probability of drawing a path from a subset $\mathcal{P}^{a} \subseteq \mathcal{P}$ knowing that we are in $\mathcal{P}^{b} \subseteq \mathcal{P}$ (with $\mathcal{P}^{a} \subseteq \mathcal{P}^{b}$), we will use
\begin{equation}
\text{P}(\mathcal{P}^{a} | \mathcal{P}^{b}) \triangleq \text{P}(\wp \in \mathcal{P}^{a} | \wp \in \mathcal{P}^{b}) = \frac{w(\mathcal{P}^{a})}{w(\mathcal{P}^{b})}.
\end{equation}
As probabilities restrained on the bag-of-hitting-paths form an important part of this work, we will often use the notation $\text{P}^\mathrm{h}(\mathcal{Q}^\mathrm{h}) \triangleq \text{P}( \mathcal{Q}^\mathrm{h} | \mathcal{P}^\mathrm{h})$ for any subset $\mathcal{Q}^\mathrm{h} \subseteq \mathcal{P}^\mathrm{h}$. In other words, we define the \emph{generalized bag-of-hitting-paths probabilities} of drawing the hitting-path $\wp^\mathrm{h}$ by
\begin{equation}
\text{P}^\mathrm{h}(\wp^\mathrm{h}) = \frac{w(\wp^\mathrm{h})}{\sum_{\tilde{\wp}^\mathrm{h} \in \mathcal{P}^\mathrm{h}} w(\tilde{\wp}^\mathrm{h})} = \frac{w(\wp^\mathrm{h})}{w(\mathcal{P}^\mathrm{h})}.
\label{Eq_hittingPathProbability01}
\end{equation}


\subsubsection{Consistency condition on the weighted adjacency matrix $\mathbf{W}$}
\label{W_radius}

As stated before, we require non-null bag-of-paths probabilities, which means a finite weight for the whole bag of paths $\mathcal{P}$. Note that the subsets of paths $\mathcal{P}_{st}$ with $s,t = 1, \dots, n$ form a partition of the bag-of-paths, which implies
\begin{equation}
w(\mathcal{P}) = \sum_{s,t \in \mathcal{V}} w(\mathcal{P}_{st}) = \sum_{s,t \in \mathcal{V}}  \sum_{\wp_{st} \in \mathcal{P}_{st}} w(\wp_{st}) = \sum_{s,t \in \mathcal{V}}  \left[\sum_{\tau=0}^\infty \mathbf{W}^{\tau} \right]_{st}.
\end{equation}
This shows that $w(\mathcal{P})$ is finite if and only if $\sum_{\tau=0}^\infty \mathbf{W}^{\tau}$ converges.
Let $\rho(\mathbf{W})$ be the \emph{spectral radius} of this weighted adjacency matrix, i.e., the largest modulus of its eigenvalues \cite{Meyer-2000},
\begin{equation}
\rho(\mathbf{W}) \triangleq \max_{\lambda \in \sigma(\mathbf{W})} | \lambda |,
\end{equation}
where $\sigma(\mathbf{W})$ is the \emph{spectrum} of $\mathbf{W}$. By the Perron-Frobenius theorem (see, e.g., \cite{Langville-2006,Meyer-2000}), the non-negativity of the components of $\mathbf{W}$ implies that the largest eigenvalue is real and positive, and thus $\lambda_1 \triangleq \rho(\mathbf{W}) \ge 0$. $\lambda_1$ is also called the \emph{Perron eigenvalue}.

The series $\sum_{\tau=0}^\infty \mathbf{W}^{\tau} = \mathbf{I} + \mathbf{W} + \mathbf{W}^2 + \cdots$ is called the \emph{Neumann series} of $\mathbf{W}$ \cite{Meyer-2000}. Concerning Neumann series, the following statements are equivalent \cite{Meyer-2000} 
\begin{enumerate}
  \item $\rho(\mathbf{W}) < 1$,
  \item $\lim_{\tau \to \infty} \mathbf{W}^{\tau} = 0$,
  \item $\sum_{\tau=0}^\infty \mathbf{W}^{\tau}$ converges.
\end{enumerate}
In this case, $(\mathbf{I} - \mathbf{W})^{-1}$ exists with $\sum_{\tau=0}^\infty \mathbf{W}^{\tau} = (\mathbf{I} - \mathbf{W})^{-1}$, and we can define $\mathbf{Z} = (z_{ij})$, the \emph{fundamental matrix} of the bag-of-paths system, as
\begin{equation}
\mathbf{Z} \triangleq (\mathbf{I} - \mathbf{W})^{-1}.
\label{eq:fundamentalMatrixDefinition01}
\end{equation}
Thus, by restricting ourselves to graphs with a weighted adjacency matrix verifying $\rho(\mathbf{W}) < 1$, we ensure that bag-of-paths probabilities are well-defined. Now, it is known that each irreducible (non-negative) substochastic matrix has this property (see \cite{Meyer-2000}, p.\ 685, exercise 8.3.7). Therefore, the property holds for strongly connected graphs with a substochastic weight matrix $\mathbf{W}$ which, instead otherwise stated, will be assumed for now. More generally and intuitively, $\rho(\mathbf{W}) < 1$ should hold provided that each node of $G$ is connected through directed links to at least one killing node -- a node whose (weighted) outdegree is strictly less than one.
Another interesting property of an irreducible substochastic weight matrix $\mathbf{W}$ is that all of its elements are smaller or equal to $1$, because its row sums are lesser or equal to $1$.

In the case of a strongly connected graph and a weight matrix smaller than $1$, several important quantities defined on subsets of the bag-of-paths can be expressed through the components of the fundamental matrix, as will be shown in Section \ref{section_weight_formulae}.


\subsection{Nodes (co-)presence and (co-)occurrences on paths}

The introduced similarity measures between nodes will be based on node presences and node occurrences on paths. We now introduce some notations related to these quantities that will be used all along the paper.
Let the \emph{presence} variable of node $i$ on a given observed path $\wp$ be
\begin{equation}
\delta(i \in \wp) \triangleq \begin{cases} 1 & \mbox{if } i \in \wp, \\ 0 & \mbox{otherwise}.
\end{cases}
\end{equation}
Moreover, let the \emph{number of occurrences}, or simply \emph{occurrences}, variable for node $i$ on path $\wp = (i_0, \dots, i_{\ell})$ be
\begin{equation}
\eta(i \in \wp) \triangleq \sum_{\tau = 0}^{\ell} \delta_{ii_\tau},
\end{equation}
where $\delta_{ii_\tau}$ is the \emph{Kronecker delta} between $i$ and $i_\tau$, i.e., $\delta_{ii_\tau}= 1$ if $i_\tau = i$ (node at position $\tau$ on path $\wp$ is equal to node $i$) and $\delta_{ii_\tau}=0$ otherwise. These variables are also used to signify the report the presence of two nodes: let \emph{co-presences} and \emph{co-occurrences} of nodes $i$ and $j$ on path $\wp$ be, respectively, $\delta(i \in \wp) \delta(j \in \wp)$ and $\eta(i \in \wp) \eta(j \in \wp)$.

In this work, we will mainly be interested in computing \emph{covariances} and \emph{correlations} of presence and occurrence variables between nodes, defined with respect to either bag-of-paths or bag-of-hitting-paths probabilities. These covariances and correlations between nodes are semi-definite positive by definition as they are inner product, or Gram, matrices \cite{Olver-2006}. They will be used in Section \ref{case_studies} as \emph{kernel matrices} in order to, e.g., perform semi-supervised classification tasks.
Note that expected values of presence and occurrences also define \emph{centrality indices}, generalizing some other centrality measures such as the \emph{betweenness centrality} \cite{Freeman-1977,Freeman-1978} or the \emph{random walk centrality} \cite{brandes2005centrality,Newman-05}. However, studying these centrality indices is out of the scope of this work, and were already investigated in \cite{kivimaki2016two}.


\subsection{Some examples of weight matrix}
\label{Subsec_examples_weight_matrix01}


\subsubsection{A particular case: the standard bag-of-paths framework}
\label{classical_BoP}

The standard bag-of-paths framework is a good example of such a weighting scheme \cite{Francoisse-2017,Lebichot-2014,Mantrach-2009}. In that context, graph $G$ is represented by its weighted adjacency matrix, $\mathbf{A} = (a_{ij})$, with no special requirement except that $a_{ij} \geq 0$ and the fact that the graph is strongly connected. This also allows us to derive a \emph{reference transition probabilities matrix} $\mathbf{P}_{\mathrm{ref}} = \mathbf{D}^{-1}\mathbf{A}$ of a natural random walk on $G$ with elements $p^{\mathrm{ref}}_{ij}$, where $\mathbf{D}$ is the diagonal matrix containing node out-degrees. Moreover, we assume a \emph{cost matrix} $\mathbf{C} = (c_{ij})$, which can be defined either independently from weights $a_{ij}$, or thanks to $c_{ij} = 1/a_{ij}$ or $c_{ij} = 1$ (among others). In this context, any observed path $\wp= (i_0,\dots,i_{\ell})$ induces a \emph{likelihood} $\pi^\mathrm{ref}(\wp)$, defined by $\pi^\mathrm{ref}(\wp) \triangleq \Pi_{\tau = 0}^{\ell-1} p^\mathrm{ref}_{i_{\tau},i_{\tau+1}}$ and a \emph{cost} $c(\wp)$ defined by $c(\wp) \triangleq \sum_{\tau = 0}^{\ell-1} c_{i_{\tau},i_{\tau+1}}$. The bag-of-paths probabilities are constructed in order to favor paths of low cost subject to a Kullback-Leibler divergence (KL) level, i.e., by solving the following problem, minimizing free energy \cite{Francoisse-2017,Mantrach-2009}:
\begin{equation}
\vline \begin{array}{ll@{}ll}
\underset{\{ \mathrm{P}(\wp) \}_{\wp \in \mathcal{P}} }{\text{minimize}}  & \displaystyle\sum\limits_{\wp \in \mathcal{P}} \mathrm{P}(\wp) {c}(\wp) + T \sum_{\wp \in \mathcal{P}} \mathrm{P}(\wp) \log(\mathrm{P}(\wp)/\mathrm{P}_\mathrm{ref}(\wp)), \\
\text{subject to}& \sum_{\wp \in \mathcal{P}} \mathrm{P}(\wp) = 1,
\end{array} \label{old_prob}
\end{equation}
where $T > 0$, the \emph{temperature} is a free parameter monitoring the KL level, and $\mathrm{P}^\mathrm{ref}(\wp)$ is the random walk reference probability of a path $\wp$ proportional to its likelihood, i.e, $\mathrm{P}^\mathrm{ref}(\wp) = {\pi}^\mathrm{ref}(\wp) / \sum_{\wp' \in \mathcal{P}}  {\pi}^\mathrm{ref}(\wp')$.

As for maximum entropy problems \cite{Cover-2006,Kapur-1989,Kapur-1992}, solving this problem yields a \emph{Gibbs-Boltzmann probability distribution} \cite{Francoisse-2017,Mantrach-2009}
\begin{equation}
\mathrm{P}^\star(\wp) = \frac{{\pi}^\mathrm{ref}(\wp) \exp(- \beta {c}(\wp)) }{\sum_{\wp' \in \mathcal{P}} {\pi}^\mathrm{ref}(\wp') \exp(- \beta {c}(\wp'))}, \label{gibbs}
\end{equation}
where $\beta \triangleq 1/T$ is the \emph{inverse temperature} parameter. This solution allows us to choose paths according to $\mathrm{P}^\mathrm{ref}(\wp)$ when the temperature $T$ is high, and increases the probability of choosing low-cost paths as the temperature decreases, up to eventually selecting shortest paths only when $T \rightarrow 0^{+}$.

Interestingly, for a path $\wp= (i_0,\dots,i_{\ell})$, the numerator in (\ref{gibbs}) can be written as ${\pi}^\mathrm{ref}(\wp) \exp(- \beta {c}(\wp))$ $=$ $\prod_{\tau = 0}^{\ell-1} p^\mathrm{ref}_{i_{\tau},i_{\tau+1}} \exp(- \beta c_{i_{\tau},i_{\tau+1}}),$ which means that it corresponds to a generalized path weight $w(\wp)$ built from the matrix $\mathbf{W}$ defined by
\begin{equation}
\mathbf{W} \triangleq \mathbf{P}_\mathrm{ref} \circ \exp[-\beta \mathbf{C}],
\label{Eq_weighted_adjacency_matrix01}
\end{equation}
where $\exp[-\beta \mathbf{C}]$ is the component-wise exponential and $\circ$ the Hadamard product. We observe that if the cost matrix has at least one non-zero value for a given edge, the matrix $\mathbf{W}$ is substochastic. Moreover, when the graph is strongly connected, we saw that the substochastic weight matrix verifies $\rho(\mathbf{W}) < 1$, which implies that the standard bag-of-paths model is indeed a particular case of the generalized one. In the case studies of Section \ref{case_studies}, this standard bag-of-paths framework will be used in the two considered applications.


\subsubsection{Another particular case: absorbing Markov chains}
\label{Subsub_absorbingMarkovChain01}

Let us now consider absorbing Markov chains. In this context, some nodes, called \emph{absorbing nodes} $\mathcal{A}$, are trapping the random walker so that the corresponding rows of the transition matrix of the Markov chain $\mathbf{P}$ are set to $0$, except on the diagonal where we find a $1$ \cite{Kemeny-1960,Snell-1984,Grinstead-1997,Norris-1997,Taylor-1998}. However, in that case, the matrix $(\mathbf{I} - \mathbf{P})$, which would be a good candidate for the fundamental matrix of the process, is not of full rank and its inverse is not defined. Then, one has to rely on pseudo-inverses which introduces an extra level of complexity to the theory of finite Markov chains \cite{Kemeny-1960,Snell-1984,Grinstead-1997}.

One simple trick solving this issue is to consider \emph{killing}, absorbing, nodes, which simply aims at setting the corresponding rows of the original transition matrix (considered as stochastic and irreducible) to $0$ \cite{Fouss-2016,Klenke-2014}. Thus, we consider that the random walker stops its walk after reaching such a killing, absorbing, node. The matrix of this killed process, denoted by $\mathbf{W}$, is then substochastic and has a spectral radius lesser than one, even if the corresponding graph is not strongly connected any more. Then, the fundamental matrix is simply $\mathbf{Z} = (\mathbf{I} - \mathbf{W})^{-1}$ and all the quantities of interest can be computed from its elements. For instance, the probability of being killed in node $t \in \mathcal{A}$ when starting a random walk from node $s$ is
\begin{equation*}
    \frac{ \sum_{\wp_{st} \in \mathcal{P}_{st}} w(\wp_{st}) } {\sum_{a \in \mathcal{A}} \sum_{\wp_{sa} \in \mathcal{P}_{sa}} w(\wp_{sa}) }
    = \frac{ z_{st} } { \sum_{a \in \mathcal{A}} z_{at} }
\end{equation*}
which relies on Result (R.1) stated later, and is quite intuitive.
Let us now turn to the computation of the various quantities of interest.


\section{Basic expressions for computing weights of various subsets of paths}
\label{section_weight_formulae}

We saw how the generalized bag-of-paths formalism revolves around computing weights associated to subsets of paths in order to compute quantities of interest. In this section, we assume a strongly connected graph as well as $\rho(\mathbf{W}) < 1$. In this case, we show that weights of various paths subsets can be computed directly from the components of the fundamental matrix $\mathbf{Z} = (\mathbf{I} - \mathbf{W})^{-1}$. When expressed properly, these results are quite intuitive, although their derivations can be rather tedious. Therefore, all proofs are reported in Appendix \ref{ App_A}.

Note that these developments only concern subsets where the starting and ending nodes are known exactly. However, the weights of subsets with undetermined starting and ending location are easily found. Indeed, sets $\mathcal{P}_{st}$ form a partition of $\mathcal{P}$, implying $w(\mathcal{P}) = \sum_{s,t \in \mathcal{V}} w(\mathcal{P}_{st})$. All subsets defined by a particular superscript can be handled in exactly the same way, e.g., $w\big(\mathcal{P}^{\mathrm{h}(-i)}\big) = \sum_{s,t \in \mathcal{V}} w\big(\mathcal{P}^{\mathrm{h}(-i)}_{st}\big)$.
The results (numbered (R.1)-(R.18)) are presented in a sequential way, from the most obvious to the most complex. Moreover, each derived quantity usually depends on the previous ones.


\begin{table}[t]
\begin{center}
\footnotesize
\begin{tabular}{|l|l|}
\hline
$w_{ij} = [\mathbf{W}]_{ij}$ & element $i,j$ of the weight matrix of graph $G$ \\\hline
$z_{st} = [\mathbf{Z}]_{st} = w\left(\mathcal{P}_{st}\right)$ & element $s,t$ of the fundamental matrix $\mathbf{Z}$ computing the total weight on regular paths \\\hline
$z^\mathrm{h}_{st} =  w\left(\mathcal{P}^\mathrm{h}_{st}\right)$ & total weight of hitting paths connecting $s$ to $t$ \\\hline
$z^{(+i)}_{st} = w\left(\mathcal{P}^{(+i)}_{st}\right)$ & total weight of regular paths connecting $s$ to $t$ and visiting intermediate node $i$ \\\hline
$z^{\mathrm{h}(+i)}_{st} = w\left(\mathcal{P}^{\mathrm{h}(+i)}_{st}\right)$ & total weight of hitting paths connecting $s$ to $t$ and visiting intermediate node $i$ \\\hline
$z^{(-i)}_{st} = w\left(\mathcal{P}^{(-i)}_{st}\right)$ & total weight of regular paths connecting $s$ to $t$ and avoiding node $i$ \\\hline
$z^{\mathrm{h}(-i)}_{st} = w\left(\mathcal{P}^{\mathrm{h}(-i)}_{st}\right)$ & total weight of hitting paths connecting $s$ to $t$ and avoiding node $i$ \\\hline
$z^{(+\{i,j\})}_{st} = w\left(\mathcal{P}^{(+\{i,j\})}_{st}\right)$ & total weight of regular paths connecting $s$ to $t$ and visiting intermediate nodes $i$, $j$ \\\hline
$z^{\mathrm{h}(+\{i,j\})}_{st} = w\left(\mathcal{P}^{\mathrm{h}(+\{i,j\})}_{st}\right)$ & total weight of hitting paths connecting $s$ to $t$ and visiting intermediate nodes $i$, $j$ \\\hline
$z^{(-\{i,j\})}_{st} = w\left(\mathcal{P}^{(-\{i,j\})}_{st}\right)$ & total weight of regular paths connecting $s$ to $t$ and avoiding nodes $i$, $j$ \\\hline
$z^{\mathrm{h}(-\{i,j\})}_{st} = w\left(\mathcal{P}^{\mathrm{h}(-\{i,j\})}_{st}\right)$ & total weight of hitting paths connecting $s$ to $t$ and avoiding nodes $i$, $j$ \\\hline

$z^{(+\mathcal{I})}_{st} = w\left(\mathcal{P}^{(+\mathcal{I})}_{st}\right)$ & total weight of regular paths connecting $s$ to $t$ and visiting intermediate nodes in set $\mathcal{I}$ \\\hline
$z^{\mathrm{h}(+\mathcal{I})}_{st} = w\left(\mathcal{P}^{\mathrm{h}(+\mathcal{I})}_{st}\right)$ & total weight of hitting paths connecting $s$ to $t$ and visiting intermediate nodes in set $\mathcal{I}$ \\\hline
$z^{(-\mathcal{I})}_{st} = w\left(\mathcal{P}^{(-\mathcal{I})}_{st}\right)$ & total weight of regular paths connecting $s$ to $t$ and avoiding nodes in set $\mathcal{I}$ \\\hline
$z^{\mathrm{h}(-\mathcal{I})}_{st} = w\left(\mathcal{P}^{\mathrm{h}(-\mathcal{I})}_{st}\right)$ & total weight of hitting paths connecting $s$ to $t$ and avoiding nodes in set $\mathcal{I}$ \\\hline
\end{tabular}
\caption{Notation for the $z$ auxiliary variables computing total weights of subsets of paths on $G$, for both regular and hitting paths.}
\label{Tab_notation02}
\end{center}
\end{table}


\subsection{Key relationships}

As a starting point, the following closed-form expressions, already known in the standard bag-of-paths and RSP frameworks \cite{Francoisse-2017,Kivimaki-2012}, are derived in Appendix \ref{ App_A} for both regular and hitting paths,
\begin{align}
z_{st} &= w\left(\mathcal{P}_{st}\right) = [\mathbf{Z}]_{ij} = \sum_{\wp \in \mathcal{P}_{st}} w(\wp), \tag{\textbf{R.1}} \\
z^\mathrm{h}_{st} &\triangleq  w\left(\mathcal{P}^\mathrm{h}_{st}\right) = \sum_{\wp \in \mathcal{P}_{st}^{\mathrm{h}}} w(\wp) = \frac{z_{st}}{z_{tt}},  \tag{\textbf{R.2}} 
\end{align}
and we observe that $z^\mathrm{h}_{tt} = 1$ from (R.2). Moreover, we have $z_{st} = z^\mathrm{h}_{st} z_{tt}$ which will be useful later. Thus, the sum of the weights of regular paths between two nodes is given by the corresponding element of the fundamental matrix (\ref{eq:fundamentalMatrixDefinition01}).

\subsection{Computing weights for sets of paths containing or avoiding nodes}

In this subsection, a number of \emph{auxiliary $z$-quantities} computing total weights of subsets of paths are defined and calculated. They serve as building blocks for computing association measures between nodes, like covariance and correlation. We start with quantities involving visits to one intermediary node, then we consider visiting two nodes, and we finally extend the results to an arbitrary number of nodes. A summary of the relevant notation appears in Table \ref{Tab_notation02}. 

\subsubsection{Path weights containing or avoiding one node}

Expressions computing the total weight for sets of paths containing or avoiding a particular node $i$ can be further developed for both regular and hitting paths (see Appendix \ref{ App_A} for derivations and details). For regular paths,
\begin{alignat}{3}
z^{(+i)}_{st} &\triangleq w\left(\mathcal{P}^{(+i)}_{st}\right) = \sum_{\wp \in \mathcal{P}_{st}} \delta(i \in \wp) \, w(\wp) =  z^\mathrm{h}_{si} z_{it},
\tag{\textbf{R.3}}
\end{alignat}
and we recall that indicator function $\delta(i \in \wp) = 1$ when path $\wp$ contains node $i$ and 0 otherwise.
It can be observed that $z^{(+i)}_{st}$ reduces to $z_{st}$ when $i=s$ and when $i=t$, which is natural.

From Equation (R.3), we further obtain, when avoiding $i$,
\begin{alignat}{3}
z^{(-i)}_{st} &\triangleq w\left(\mathcal{P}^{(-i)}_{st}\right) = \sum_{\wp \in \mathcal{P}_{st}} (1 - \delta(i \in \wp)) \, w(\wp) \nonumber \\
& = z_{st} - z^{(+i)}_{st} = z_{st} - z^\mathrm{h}_{si} z_{it},
\tag{\textbf{R.4}}
\end{alignat}
and this time,  by (R.2), $z^{(-i)}_{st} = 0$ when $i=s$ and when $i=t$, as should be. For hitting paths now,
\begin{align}
z^{\mathrm{h}(-i)}_{st} &\triangleq w\left(\mathcal{P}^{\mathrm{h}(-i)}_{st}\right) = \sum_{\wp \in \mathcal{P}^{\mathrm{h}}_{st}} (1 - \delta(i \in \wp)) \, w(\wp) \notag \\
&=
    \begin{cases}
		\dfrac{ z^{(-i)}_{st} }{ z^{(-i)}_{tt} } \\
		0
    \end{cases}
=
    \begin{cases}
		\dfrac{z^\mathrm{h}_{st} - z^\mathrm{h}_{si} z^\mathrm{h}_{it}}{1 - z^\mathrm{h}_{ti} z^\mathrm{h}_{it}}& \mbox{if } i\neq t, \\
		0 & \mbox{if } i = t,
    \end{cases}
\tag{\textbf{R.5}}
\end{align}
where $z^{\mathrm{h}(-i)}_{st} = 0$ when $i=s$, as should be. Observe the similarity with (R.2). Note that when $s=t$, the result is equal to 1.
Considering now the presence of node $i$,
\begin{equation}
z^{\mathrm{h}(+i)}_{st} \triangleq w\left(\mathcal{P}^{\mathrm{h}(+i)}_{st}\right) = \sum_{\wp \in \mathcal{P}^{\mathrm{h}}_{st}} \delta(i \in \wp) \, w(\wp) = 
    \begin{cases}
		z^{\mathrm{h}(-t)}_{si} z^\mathrm{h}_{it} & \mbox{if } i\neq t, \\
		z^\mathrm{h}_{st} & \mbox{if } i = t,
	\end{cases}
\tag{\textbf{R.6}}
\end{equation}
and we naturally obtain $z^{\mathrm{h}(+i)}_{st} = z^{\mathrm{h}}_{st}$ when $i=s$.

\subsubsection{Path weights containing or avoiding two nodes}

In this subsection, we consider that $i \neq j$ and the expressions are only valid in this situation. When this is not the case, the problem is reduced to the task of finding only one node on paths, and its solution is given in the previous section with, e.g, $w\big(\mathcal{P}^{(+\{i,i\})}_{st}\big) = w\big(\mathcal{P}^{(+i)}_{st}\big)$.
In Appendix \ref{ App_A}, the expressions computing the total weight on sets of paths containing or avoiding two nodes of interest $i$ and $j$, for both regular and hitting paths, are derived. The main results are summarized in (R.7)-(R.10),
\begin{align}
z^{(+\{i,j\})}_{st} &\triangleq w\left(\mathcal{P}^{(+\{i,j\})}_{st}\right) = \sum_{\wp \in \mathcal{P}_{st}} \delta(i \in \wp)\delta(j \in \wp) \, w(\wp) \notag \\
&= z^{\mathrm{h}(+i)}_{sj} z_{jt} + z^{\mathrm{h}(+j)}_{si} z_{it} \qquad\qquad \mbox{if } i \ne j. \tag{\textbf{R.7}} 
\end{align}
Notice that this expression is coherent when $i=s$, $j=s$, $i=t$ and $j=t$ as it reduces to the expressions involving only one node (R.3).
Three different expressions (R.8.1)-(R.8.3) can be derived for computing the next quantity,
\begin{alignat}{3}
z^{(-\{i,j\})}_{st} &\triangleq w\left(\mathcal{P}^{(-\{i,j\})}_{st}\right) = \sum_{\wp \in \mathcal{P}_{st}} (1 - \delta(i \in \wp)) && (1 - \delta(j \in \wp)) \, w(\wp) \notag \\
&= z_{st} - z^{\mathrm{h}(-j)}_{si} z_{it}  - z^{\mathrm{h}(-i)}_{sj} z_{jt} && \mbox{if } i \ne j \tag{\textbf{R.8.1}} \\
&= z^{(-j)}_{st} - z^{\mathrm{h}(-j)}_{si} z^{(-j)}_{it} && \mbox{if } i \ne j \tag{\textbf{R.8.2}} \\
&= z^{(-i)}_{st} - z^{\mathrm{h}(-i)}_{sj} z^{(-i)}_{jt} && \mbox{if } i \ne j. \tag{\textbf{R.8.3}}
\end{alignat}
The following results (R.9) and (R.10) for hitting paths assume that $i \neq j \neq t$. If either $i = t$ or $j = t$, the quantity (R.9) must be equal to 0 (the destination node $t$ cannot be avoided). When $i=j$ (only one node is present or absent), the Equations (R.5) and (R.6) must be used instead.
Once again, three alternative expressions (R.9.1)-(R.9.3) can be derived from the more general result (R.9) for computing the equivalent quantity for hitting paths,
\begin{alignat}{3}
z^{\mathrm{h}(-\{i,j\})}_{st} &\triangleq w\left(\mathcal{P}^{\mathrm{h}(-\{i,j\})}_{st}\right)
= \sum_{\wp \in \mathcal{P}^{\mathrm{h}}_{st}} (1 - \delta(i &&\in \wp)) (1 - \delta(j \in \wp)) \, w(\wp) \notag \\
 &= \dfrac{ z^{(-\{i,j\})}_{st} }{ z^{(-\{i,j\})}_{tt} } && \mbox{if } i,j \neq t \land i \ne j  \tag{\textbf{R.9}} \\
 &= \frac{z^\mathrm{h}_{st} - z^{\mathrm{h}(-j)}_{si} z^\mathrm{h}_{it}  - z^{\mathrm{h}(-i)}_{sj} z^\mathrm{h}_{jt} }{1 - z^{\mathrm{h}(-j)}_{ti} z^\mathrm{h}_{it}  - z^{\mathrm{h}(-i)}_{tj} z^\mathrm{h}_{jt} } && \mbox{if } i,j \neq t \land i \ne j \tag{\textbf{R.9.1}} \\
 &= \frac{z^{\mathrm{h}(-j)}_{st} - z^{\mathrm{h}(-j)}_{si} z^{\mathrm{h}(-j)}_{it}}{1 - z^{\mathrm{h}(-j)}_{ti} z^{\mathrm{h}(-j)}_{it}} && \mbox{if } i,j \neq t \land i \ne j \tag{\textbf{R.9.2}} \\
 &= \frac{z^{\mathrm{h}(-i)}_{st} - z^{\mathrm{h}(-i)}_{sj} z^{\mathrm{h}(-i)}_{jt}}{1 - z^{\mathrm{h}(-i)}_{tj} z^{\mathrm{h}(-i)}_{jt}} && \mbox{if } i,j \neq t \land i \ne j \tag{\textbf{R.9.3}} \\
&= 0. && \mbox{if } i = t \lor j = t,  \tag{\textbf{R.9}}
\end{alignat}
Note also that the result is $1$ when $s=t$.
Finally, for node presence, we obtain
\begin{alignat}{3}
z^{\mathrm{h}(+\{i,j\})}_{st} &\triangleq w\left(\mathcal{P}^{\mathrm{h}(+\{i,j\})}_{st}\right) = \sum_{\wp \in \mathcal{P}^{\mathrm{h}}_{st}} \delta(i \in \wp) \delta(j \in \wp) &&\, w(\wp) \notag \\
&= z^{\mathrm{h}(-\{j,t\})}_{si} z^{\mathrm{h}(+j)}_{it} + z^{\mathrm{h}(-\{i,t\})}_{sj} z^{\mathrm{h}(+i)}_{jt} && \mbox{if } i,j \neq t \land i \ne j \tag{\textbf{R.10}}
\end{alignat}
and, again, this expression is coherent when $i=s$, $j=s$, $i=t$ and $j=t$ as it reduces to (R.6). When $s=t$, the result is simply $0$.

\subsubsection{Path weights containing or avoiding sets of nodes}

Up to now, we were mainly interested in computing weights on subsets of paths containing or avoiding one or two nodes. However, it is possible to deal with subsets with higher numbers of nodes through some recurrence formulae.

Let us assume we have a set of distinct nodes $\mathcal{I}$, where $s,t \notin \mathcal{I}$, and let $\mathcal{S}$ be a subset of $\mathcal{I}$, i.e., $\mathcal{S} \subset \mathcal{I}$. The weight of the regular paths avoiding all nodes in $\mathcal{S}$ is denoted as $z^{(-\mathcal{S})}_{st} = w\big(\mathcal{P}^{(-\mathcal{S})}_{st}\big)$ and the corresponding weight for hitting paths by $z^{\mathrm{h}(-\mathcal{S})}_{st} = w\big(\mathcal{P}^{\mathrm{h}(-\mathcal{S})}_{st}\big)$. The same convention, but with a $+$ sign this time, is used for paths containing a set of nodes $\mathcal{S}$ (all nodes in $\mathcal{S}$ present). We would like to compute quantities like $z^{(+\mathcal{I})}_{st}$, $z^{\mathrm{h}(+\mathcal{I})}_{st}$, $z^{(-\mathcal{I})}_{st}$ and $z^{\mathrm{h}(-\mathcal{I})}_{st}$ for the set $\mathcal{I}$ in function of some subsets $\mathcal{S}_{i}$ of $\mathcal{I}$.

First, let us compute the weights of avoiding paths. Let $\mathcal{S}_i \triangleq \mathcal{I} \setminus i, \, \forall i \in \mathcal{I}$, meaning that $\mathcal{I} = \mathcal{S}_i \cup \{ i \}$. We obtain (see Appendix \ref{ App_A})
\begin{alignat}{3}
&z^{(-\mathcal{I})}_{st} &&\triangleq w\left(\mathcal{P}^{(-\mathcal{I})}_{st}\right) = z^{(-\mathcal{S}_i)}_{st} - z^{\mathrm{h}(-\mathcal{S}_i)}_{si} z^{(-\mathcal{S}_i)}_{it}  &\quad \forall i \in \mathcal{I},  \tag{\textbf{R.11}}\\ 
&z^{\mathrm{h}(-\mathcal{I})}_{st} &&\triangleq w\left(\mathcal{P}^{\mathrm{h}(-\mathcal{I})}_{st}\right) = \frac{ z^{(-\mathcal{I})}_{st} } { z^{(-\mathcal{I})}_{tt} } = \frac{ z^{\mathrm{h}(-\mathcal{S}_i)}_{st} - z^{\mathrm{h}(-\mathcal{S}_i)}_{si} z^{\mathrm{h}(-\mathcal{S}_i)}_{it}}{1 - z^{\mathrm{h}(-\mathcal{S}_i)}_{ti}z^{\mathrm{h}(-\mathcal{S}_i)}_{it}}   &\quad \forall i \in \mathcal{I}.  \tag{\textbf{R.12}}
\end{alignat}

Conversely, weights on sets of paths containing some predefined nodes in $\mathcal{I}$ can be obtained  (see Appendix \ref{ App_A}) from the previously computed quantities (R.11) and (R.12) thanks to
\begin{alignat}{1}
&z^{(+\mathcal{I})}_{st} \triangleq w\left(\mathcal{P}^{(+\mathcal{I})}_{st}\right) = z_{st} + \sum_{\mathcal{S} \subseteq \mathcal{I} \atop \mathcal{S} \ne \varnothing} (-1)^{|\mathcal{S}|} z^{(-\mathcal{S})}_{st} = \sum_{\mathcal{S} \subseteq \mathcal{I}} (-1)^{|\mathcal{S}|} z^{(-\mathcal{S})}_{st}, \tag{\textbf{R.13}} \\ 
&z^{\mathrm{h}(+\mathcal{I})}_{st} \triangleq w\left(\mathcal{P}^{\mathrm{h}(+\mathcal{I})}_{st}\right) = z^\mathrm{h}_{st} + \sum_{\mathcal{S} \subseteq \mathcal{I} \atop \mathcal{S} \ne \varnothing} (-1)^{|\mathcal{S}|} z^{\mathrm{h}(-\mathcal{S})}_{st} = \sum_{\mathcal{S} \subseteq \mathcal{I}} (-1)^{|\mathcal{S}|} z^{\mathrm{h}(-\mathcal{S})}_{st}, \tag{\textbf{R.14}}
\end{alignat}
where the summation on $\mathcal{S} \subseteq \mathcal{I}$ with $\mathcal{S} \ne \varnothing$ means a summation on all subsets $\mathcal{S}$ of $\mathcal{I}$, except the empty set. Moreover, by convention, $(-1)^{0}=1$.

Let us take an example with set $\mathcal{I} = \{ i,j \}$ and regular paths. We easily obtain from (R.13)
\begin{equation*}
z^{(+\{ i,j \})}_{st} = z_{st} - z^{(-i)}_{st} - z^{(-j)}_{st} + z^{(-\{ i,j \})}_{st},
\end{equation*}
and $z^{(-\{ i,j \})}_{st}$ is computed with (R.11), $z^{(-\{ i,j \})}_{st} = z^{(-j)}_{st} - z^{\mathrm{h}(-j)}_{si} z^{(-j)}_{it}$, which corresponds to (R.8.2). It can be easily verified numerically that the obtained expression for $z^{(+\{ i,j \})}_{st}$ provides the same results as (R.7).

We now turn to the computation of related quantities involving, this time, the number of visits (instead of presence) to some set of predefined nodes.


\subsection{Computing weights of node (co)-occurrences on sets of paths}
\label{section_pass_formulae}

This section will derive equivalent results for the number of occurrences of nodes (and not simply the presence of the node as in the previous section) on paths of the graph.
The first moments of occurrence variables of the type $\eta(i \in \wp)$, i.e., the number of times node $i$ is visited on path $\wp$, can be obtained, but in a very different way than in the previous subsections. Indeed, most of the results of this subsection are calculated by taking some partial derivatives with respect to path weights, as shown in Appendix \ref{App_B} and already exploited in \cite{Mantrach-2009} for the standard bag-of-paths framework and regular paths. Note that we are not interested in paths avoiding nodes in this section as they correspond to paths with zero presence, which was covered in the previous section.

We now compute the weighted number of occurrences of some nodes on the sets of regular paths $\mathcal{P}$ and hitting paths $\mathcal{P}^{\mathrm{h}}$, expressed in function of the fundamental matrix. This provides (see Appendix \ref{App_B} for details), for regular paths,
\begin{equation}
\sum_{\wp_{st} \in \mathcal{P}_{st}} \eta(i \in \wp_{st}) \, w(\wp_{st}) = z_{si} z_{it}, \tag{\textbf{R.15}} \\
\end{equation}

\begin{equation}
\sum_{\wp_{st} \in \mathcal{P}_{st}} \eta(i \in \wp_{st}) \eta(j \in \wp_{st}) \, w(\wp_{st}) \nonumber \\
= z_{si}z_{ij}z_{jt} + z_{sj}z_{ji}z_{it} - \delta_{ij} z_{si} z_{jt}. \tag{\textbf{R.16}}
\end{equation}
Note that this expression includes the $i=j$ case.
For hitting paths, we have
\begin{equation}
\sum_{\wp^\mathrm{h}_{st} \in \mathcal{P}^\mathrm{h}_{st}} \eta(i \in \wp^\mathrm{h}_{st}) \, w(\wp^\mathrm{h}_{st}) = z^{(-t)}_{si} z^\mathrm{h}_{it} + \delta_{it} z^\mathrm{h}_{st}, \tag{\textbf{R.17}}
\end{equation}
and this quantity is equal to $z^\mathrm{h}_{st}$ when $i=t$. Moreover, for two nodes,
\begin{align}
&\sum_{\wp^\mathrm{h}_{st} \in \mathcal{P}^\mathrm{h}_{st}} \eta(i \in \wp^\mathrm{h}_{st}) \eta(j \in \wp^\mathrm{h}_{st}) \, w(\wp^\mathrm{h}_{st})
= z^{(-t)}_{si}z^{(-t)}_{ij}z^\mathrm{h}_{jt} + z^{(-t)}_{sj}z^{(-t)}_{ji} z^\mathrm{h}_{it} \notag \\
& \qquad - \delta_{ij} z^{(-t)}_{si} z^\mathrm{h}_{jt} + \delta_{jt} z^{(-t)}_{si} z^\mathrm{h}_{it} + \delta_{it} z^{(-t)}_{sj} z^\mathrm{h}_{jt} + \delta_{it} \delta_{jt} z^\mathrm{h}_{st}. \tag{\textbf{R.18}}
\end{align}
These various quantities will now be used in order to define centrality and association measures on nodes.


\section{Betweenness and association measures}
\label{copre_cooc_formulae}

In this section, we will be interested in computing the expected value of (co-)presence as well as number of (co-)occurrences of nodes on paths (moments and co-moments), with respect to the generalized bag-of-paths and bag-of-hitting-paths probabilities. These quantities will allow us to compute the covariance and the correlation between node presence and occurrences. It is also shown that distance measures, extending the free energy distance \cite{Kivimaki-2012,Francoisse-2017}, can be derived from the quantities introduced so far. Note that still other similarity measures could be computed from the same expressions, like cosine measure, Jaccard index, etc.

First, let us calculate the weights of the set of all paths $\mathcal{P}$ and the set of all hitting paths $\mathcal{P}^\mathrm{h}$, as these will be used as normalizing factors in order to compute our probabilities. We readily obtain from (R.1) and (R.2)
\begin{equation}
w(\mathcal{P}) = \sum_{s,t \in \mathcal{V}} w(\mathcal{P}_{st}) = z_{\bullet \bullet}, \qquad w(\mathcal{P}^\mathrm{h}) = \sum_{s,t \in \mathcal{V}} w(\mathcal{P}^\mathrm{h}_{st}) = z^\mathrm{h}_{\bullet \bullet},
\end{equation}
where $\bullet$ indicates the summation over the corresponding index.

Then, recall that the probabilities of choosing a particular, regular, path $\wp$ (see Equation (\ref{Eq_pathProbability01})) or a hitting path $\wp^\mathrm{h}$ (see Equation (\ref{Eq_hittingPathProbability01})) are simply
\begin{equation}
\mathrm{P}(\wp) = \frac{ w(\wp) } {w(\mathcal{P})}, \qquad \mathrm{P}^\mathrm{h}(\wp) = \frac{ w(\wp^\mathrm{h}) } {w(\mathcal{P}^\mathrm{h})}.
\end{equation}
Let us now compute the quantities of interest.

\subsection{Covariance and correlation for node presence}

From (R.1), (R.2), (R.3), (R.6), the expectation of node presence, regarding $\mathrm{P}(\wp)$ (regular paths framework) and $\mathrm{P}^\mathrm{h}(\wp^\mathrm{h})$ (hitting paths framework), is easily obtained thanks to
\begin{align}
\mathbb{E}[\delta(i \in \wp)] &= \sum_{\wp \in \mathcal{P}} \delta(i \in \wp) \, \mathrm{P}(\wp)
= \sum_{\wp \in \mathcal{P}} \delta(i \in \wp) \frac{ w(\wp) } {w(\mathcal{P})} \nonumber \\
&= \frac{w\left(\mathcal{P}^{(+i)}\right)}{w(\mathcal{P})} = \frac{\sum_{s,t \in \mathcal{V}} w\left(\mathcal{P}^{(+i)}_{st}\right)}{\sum_{s,t \in \mathcal{V}} w\left(\mathcal{P}_{st}\right)} = \frac{z^\mathrm{h}_{\bullet i} z_{i\bullet}}{z_{\bullet \bullet}},
\label{Eq_cov_cor01}
\end{align}
\begin{align}
\mathbb{E}^\mathrm{h}[\delta(i \in \wp^\mathrm{h})] &= \sum_{\wp^\mathrm{h} \in \mathcal{P}^\mathrm{h}} \delta(i \in \wp^\mathrm{h}) \, \mathrm{P}(\wp^\mathrm{h})
= \frac{w\left(\mathcal{P}^{\mathrm{h}(+i)}\right)}{w(\mathcal{P^\mathrm{h}})} \nonumber \\
&= \frac{\sum_{\substack{t \in \mathcal{V}\\ t \neq i}} \big( z^{\mathrm{h}(-t)}_{\bullet i} z^\mathrm{h}_{it} \big) + z^\mathrm{h}_{\bullet i}}{{z^\mathrm{h}_{\bullet \bullet}} } 
\end{align}
where the last term is the contribution for $t=i$ as given in (R.6).

These two quantities define \emph{betweenness measures} based on node presence, quantifying to which extend each node is an important intermediary with respect to the communication (along paths) between pairs of nodes \cite{kivimaki2016two}. Communication along short paths is usually promoted by putting more weight on them, as in the standard bag of paths.

Similarly, from (R.7) and (R.10), the expected values of co-presence are
\begin{align}
\mathbb{E}[\delta(i \in \wp) \delta(j \in \wp)] &= \sum_{\wp \in \mathcal{P}} \delta(i \in \wp) \delta(j \in \wp) \, \mathrm{P}(\wp) = \frac{w\left(\mathcal{P}^{(+\{i,j\})}\right)}{w(\mathcal{P})}  \notag \\
&= \frac{z^{\mathrm{h}(+i)}_{\bullet j} z_{j\bullet} + z^{\mathrm{h}(+j)}_{\bullet i} z_{i\bullet} - \delta_{ij} z^\mathrm{h}_{\bullet i} z_{i\bullet}}{z_{\bullet \bullet}},
\label{Eq_second_moment_regular01}
\end{align}
\begin{align}
\mathbb{E}^\mathrm{h}[\delta(i \in \wp^\mathrm{h})\delta(j \in \wp^\mathrm{h})] &= \sum_{\wp^\mathrm{h} \in \mathcal{P}^\mathrm{h}} \delta(i \in \wp^\mathrm{h}) \delta(j \in \wp^\mathrm{h}) \, \mathrm{P}(\wp^\mathrm{h}) = \frac{w\left(\mathcal{P}^{\mathrm{h}(+\{i,j\})}\right)}{w(\mathcal{P}^\mathrm{h})} \notag \\
&=  \frac{\sum_{\substack{t \in \mathcal{V}\\ t \neq i,j}} \left(z^{\mathrm{h}(-\{j,t\})}_{\bullet i} z^{\mathrm{h}(+j)}_{it} + z^{\mathrm{h}(-\{i,t\})}_{\bullet j} z^{\mathrm{h}(+i)}_{jt} -  \delta_{ij} z^{\mathrm{h}(-t)}_{\bullet i} z^\mathrm{h}_{it}\right)}{z^\mathrm{h}_{\bullet \bullet}} \notag \\
&\quad + \frac{z^{\mathrm{h}(-j)}_{\bullet i} z^\mathrm{h}_{ij} + z^{\mathrm{h}(-i)}_{\bullet j} z^\mathrm{h}_{ji} + \delta_{ij} z^\mathrm{h}_{\bullet i}}{z^\mathrm{h}_{\bullet \bullet}},
\label{Eq_second_moment_hitting01}
\end{align}
where terms like $-\delta_{ij} z^\mathrm{h}_{\bullet i} z_{i\bullet}$, $\delta_{ij} z^{\mathrm{h}(-t)}_{\bullet i} z^\mathrm{h}_{it}$ and $\delta_{ij} z^\mathrm{h}_{\bullet i}$ were added in order to take into account special cases where $i = j$. Indeed, (R.7) and (R.10) concern only the situation where $i \neq j$ and must be extended in order to compute these diagonal terms.

In the first case (\ref{Eq_second_moment_regular01}), i.e., regular paths, when $i=j$, the expression in (R.7) provides $2 z^{\mathrm{h}(+i)}_{si} z_{it}$. However, the result should instead provide Equation (R.3), that is, the expression for the total weight computed on paths containing only one node $i$. But this expression (R.3) is equal to $z^{\mathrm{h}(+i)}_{si} z_{it}$ so that we have to remove one time this quantity to (R.7) in order to obtain (R.3) when $i=j$. Hence, the subtraction of $z^{\mathrm{h}(+i)}_{si} z_{it}$ appearing in Equation (\ref{Eq_second_moment_regular01}).

For the second case (\ref{Eq_second_moment_hitting01}), i.e. hitting paths, the expression in (R.10) provides $0$ when $i=j$ (and $i \ne t$). However, the result should instead be Equation (R.6), the corresponding expression for the total weight computed on hitting paths containing only one node $i$. This expression (R.6) is equal to $z^{\mathrm{h}(-t)}_{si} z^{\mathrm{h}}_{it}$ so that we have to add this quantity to (R.10) in order to obtain (R.6) when $i=j$ (and $i \ne t$). Hence, the addition of $z^{\mathrm{h}(-t)}_{si} z^{\mathrm{h}}_{it}$ appearing in the first line of Equation (\ref{Eq_second_moment_hitting01}).

The last line of Equation (\ref{Eq_second_moment_hitting01}) aims at taking care of the special cases $i=t$ and $j=t$, which are also prohibited in (R.10). In the situation $i=t$, $i$ is automatically part of the paths and the quantity $z^{\mathrm{h}(+\{i,j\})}_{st}$ should reduce to $z^{\mathrm{h}(+\{t,j\})}_{st} = z^{\mathrm{h}(+j)}_{st} = z^{\mathrm{h}(+j)}_{si}$, that is, to (R.6 with $t \ne i$). Hence the addition of the term $z^{\mathrm{h}(-i)}_{s j} z^\mathrm{h}_{ji}$ in the last line of (\ref{Eq_second_moment_hitting01}). Symmetrically, the same holds for the case $j=t$, with the introduction of $z^{\mathrm{h}(-j)}_{\bullet i} z^\mathrm{h}_{ij}$ in the last line of  (\ref{Eq_second_moment_hitting01}).

Finally, as both $z^{\mathrm{h}(-i)}_{s j} z^\mathrm{h}_{ji}$ and $z^{\mathrm{h}(-j)}_{\bullet i} z^\mathrm{h}_{ij}$ are equal to zero when $i=j$, we still need to handle the case $i=j=t$ and add the corresponding contribution. In that situation, the contribution is $z^{\mathrm{h}(+j)}_{si} = z^{\mathrm{h}(+i)}_{si}$, that is, (R.6 with $t=i$) which provides $z^\mathrm{h}_{si}$. This corresponds to the last term of Equation (\ref{Eq_second_moment_hitting01}).

These expected values are the building blocks needed to compute the \emph{covariance} and the \emph{correlation} measures between the common presence of two nodes on paths (the random variables $\delta(i \in \wp)$ and $\delta(j \in \wp)$). This provides, for regular paths,
\begin{align}
\mathrm{cov}(\delta(i \in \wp),\delta(j \in \wp)) &= \mathbb{E}[\delta(i \in \wp)\delta(j \in \wp)] - \mathbb{E}[\delta(i \in \wp)] \mathbb{E}[\delta(j \in \wp)], \label{Eq_presence_kernels01} \\
\mathrm{cor}(\delta(i \in \wp),\delta(j \in \wp)) &= \frac{\mathrm{cov}(\delta(i \in \wp),\delta(j \in \wp))}{\sqrt{\mathrm{cov}(\delta(i \in \wp),\delta(i \in \wp)) \, \mathrm{cov}(\delta(j \in \wp),\delta(j \in \wp))}},
\label{Eq_presence_kernels02}
\end{align}
and the expressions for hitting-paths probabilities are similar, with $\wp$ replaced by $\wp^{\mathrm{h}}$. The intuition behind these quantities is that two nodes are correlated when they often appear on the same, preferably short, paths.

\subsection{Covariance and correlation for the number of occurrences of nodes}

We now derive the same quantities for the number of occurrences of nodes on paths. From (R.15) and (R.17), we have for regular paths
\begin{alignat}{2}
&\mathbb{E}[\eta(i \in \wp)] &&= \sum_{\wp \in \mathcal{P}} \eta(i \in \wp) \mathrm{P}(\wp)  = \sum_{\wp \in \mathcal{P}} \eta(i \in \wp) \frac{w(\wp)}{w(\mathcal{P})} \notag \\
& &&= \sum_{s,t \in \mathcal{V}}  \sum_{\wp_{st} \in \mathcal{P}_{st}}  \eta(i \in \wp_{st}) \frac{ w(\wp_{st}) }{w(\mathcal{P})} = \frac{z_{\bullet i} z_{i \bullet}}{z_{\bullet \bullet}},
\label{Eq_first_moment_occurences01}
\end{alignat}
\begin{align}
&\mathbb{E}^\mathrm{h}[\eta(i \in \wp^\mathrm{h})] = \sum_{s,t \in \mathcal{V}}  \sum_{\wp^\mathrm{h}_{st} \in \mathcal{P}^\mathrm{h}_{st}}  \eta(i \in \wp^\mathrm{h}_{st}) \frac{w(\wp^\mathrm{h}_{st})}{w(\mathcal{P}^\mathrm{h})}
= \frac{\sum_{t \neq i} z^{(-t)}_{\bullet i} z^\mathrm{h}_{it} + z^\mathrm{h}_{\bullet i}}{z^\mathrm{h}_{\bullet \bullet}}.
\end{align}

As before, these two quantities define \emph{betweenness measures} based on node occurrences.

For the expected values of co-occurrences, (R.16) and (R.18) provide
\begin{align}
\mathbb{E}[\eta(i \in \wp)\eta(j \in \wp)] &= \sum_{s,t \in \mathcal{V}}  \sum_{\wp_{st} \in \mathcal{P}_{st}}  \eta(i \in \wp_{st}) \eta(j \in \wp_{st}) \frac{w(\wp_{st})}{w(\mathcal{P})}  \notag \\ 
&= \frac{z_{\bullet i} z_{ij} z_{j \bullet} + z_{\bullet j} z_{ji} z_{i \bullet} - \delta_{ij} z_{\bullet i} z_{j \bullet}}{z_{\bullet \bullet}},
\end{align}
\begin{align}
\mathbb{E}^\mathrm{h}[\eta(i \in \wp^\mathrm{h})\eta(j \in \wp^\mathrm{h})] &= \sum_{s,t \in \mathcal{V}}  \sum_{\wp^\mathrm{h}_{st} \in \mathcal{P}^\mathrm{h}_{st}}  \eta(i \in \wp^\mathrm{h}_{st}) \eta(j \in \wp^\mathrm{h}_{st}) \frac{w(\wp^\mathrm{h}_{st}) }{w(\mathcal{P}^\mathrm{h})} \notag \\
&= \frac{ \sum_{\substack{t \in \mathcal{V}\\ t \neq i,j}} \left( z^{(-t)}_{\bullet i}z^{(-t)}_{ij} z^\mathrm{h}_{jt} + z^{(-t)}_{\bullet j}z^{(-t)}_{ji} z^\mathrm{h}_{it} - \delta_{ij} z^{(-t)}_{\bullet i} z^\mathrm{h}_{it}\right)}{ z^\mathrm{h}_{\bullet \bullet} } \notag \\
 &\quad + \frac{z^{(-j)}_{\bullet i} z^\mathrm{h}_{ij} + z^{(-i)}_{\bullet j} z^\mathrm{h}_{ji} + \delta_{ij} z^\mathrm{h}_{\bullet i}}{ z^\mathrm{h}_{\bullet \bullet} },
 \label{Eq_second_moment_occurences01}
\end{align}
which, again, allows us to compute the covariance and the correlation, but now for number of co-occurrences of nodes on paths, with
\begin{align}
\mathrm{cov}(\eta(i \in \wp),\eta(j \in \wp)) &= \mathbb{E}[\eta(i \in \wp)\eta(j \in \wp)] - \mathbb{E}[\eta(i \in \wp)) \mathbb{E}(\eta(j \in \wp)], \label{Eq_occurences_kernels01} \\
\mathrm{cor}(\eta(i \in \wp),\eta(j \in \wp)) &= \frac{\mathrm{cov}(\eta(i \in \wp),\eta(j \in \wp))}{\sqrt{\mathrm{cov}(\eta(i \in \wp),\eta(i \in \wp)) \, \mathrm{cov}(\eta(j \in \wp),\eta(j \in \wp))}}.
\label{Eq_occurences_kernels02}
\end{align}
The expressions considering hitting-paths probabilities are similar. It is well known that such covariance and correlation matrices are positive semi-definite (Gram matrices \cite{Olver-2006}) and are therefore valid \emph{kernels on a graph} representing similarities between nodes (see \cite{Gaertner-2008,scholkopf2001learning,Shawe-Taylor-2004} for kernels in general and, e.g., \cite{FoussKernelNN-2011,Fouss-2016} for kernels on a graph).

Note that for all these formulae, computing the results for hitting paths requires an iteration over nodes $t$, which is more computationally intensive than for the non-hitting case. However, in the experiments, covariance and correlation coefficients derived from hitting paths usually performed better than their non-hitting counterparts.

\subsection{A distance measure between nodes}

When the elements of the matrix $\mathbf{W}$ represent local affinities between linked nodes, a useful distance measure -- that was called the free energy potential distance in the standard bag-of-paths framework \cite{Kivimaki-2012,Francoisse-2017} -- can be computed from previous result (R.6). The directed distance is defined as
\begin{equation}
\phi(i,j) = - \log \big( z^{\mathrm{h}}_{st} \big).
\end{equation}
It was shown in \cite{Francoisse-2017} that this quantity can be interpreted as minus $\log$ the probability of surviving during a killed random walk from $i$ to $j$, for a random walker moving according to the sub-stochastic transition matrix $\mathbf{W}$.
Then, the \emph{bag-of-paths distance} measure can immediately be deduced from the previous quantity,
\begin{equation}
\dist^{\textsc{b}o\textsc{p}}_{ij} \triangleq
\begin{cases}
\tfrac{1}{2} \big( \phi(i,j) + \phi(j,i) \big) & \text{if } i \ne j, \\
 0 & \text{if } i=j.
\end{cases}
\end{equation}
The quantity is non-negative because it represents probabilities. Moreover, it can easily be shown that the triangle inequality is satisfied. Indeed, from result (R.3), $ z^{\mathrm{h}}_{st} = z_{st} / z_{tt} \ge z^{(+i)}_{st} / z_{tt} $ (because $\mathcal{P}^{(+i)}_{st} \subseteq \mathcal{P}_{st} $). Then, from (R.3) and (R.2), the quantity on the right side of the inequality becomes $ z^{(+i)}_{st} / z_{tt} = z^{\mathrm{h}}_{si} z_{it} / z_{tt} = z^{\mathrm{h}}_{si} z^{\mathrm{h}}_{it}$. Therefore, $ z^{\mathrm{h}}_{st} \ge z^{\mathrm{h}}_{si} z^{\mathrm{h}}_{it} $ and taking $- \log$ of this inequality proves the triangle inequality for the directed distance, and thus also for the distance. In the standard bag-of-paths formalism, this distance provided good results in semi-supervised and clustering tasks \cite{Francoisse-2017,Sommer-2016,Sommer-2017}.


\section{Case study: application to semi-supervised classification}
\label{case_studies}

For illustration, the accuracy of the introduced methods will be compared on semi-supervised classification tasks. However, we have to stress that our goal here is not to propose new graph-based semi-supervised classification algorithms outperforming state-of-the-art techniques. Rather, the aim is to investigate if the introduced similarity measures are able to capture the community structure of networks in an accurate way, compared to other state-of-the-art dissimilarity measures between nodes. In our case, the main baseline method will be the free energy potential distance based on the bag-of-paths framework which performed best in a number of pattern recognition tasks. Thus, the experiments will tell us if the introduced similarity measures are competitive with respect to this free energy distance.
But before going into the details of the experiments, let us first introduce the different similarity measures that are derived from the studied models.

Following the discussion in the introduction (Subsection \ref{Subsec_intuition_models01}), our introduced measures can also be interpreted as inner products in the (usually infinite-dimensional) vector space of paths in which each node $x \in \mathcal{V}$ has a coordinate $\phi_{i}(x) = \phi(x, \wp_{i})$, for paths $\wp_{i} \in \mathcal{P}$ which have been numbered as in Section \ref{Subsec_intuition_models01}, $(\wp_{1},\wp_{2},\dots)$ by decreasing weight, so that $\mathcal{P}$ becomes a totally ordered set. For two arbitrary nodes $x, y \in \mathcal{V}$, $\bra{x}\ket{y} = \sum_{i = 1}^{\infty} \phi_{i}(x) \mathrm{P}(\wp_{i}) \phi_{i}(y)$ where $\phi_{i}(x)$ is a function of node $x$ and path $\wp_{i}$; in our case, the presence of node $x$ on path $\wp_{i} \in \mathcal{P}$ ($\phi_{i}(x) = \delta(x \in \wp_{i})$) or the number of occurrences of node $x$ on $\wp_{i}$ ($\phi_{i}(x) = \eta(x \in \wp_{i})$).

Using this property, we can define 8 different \emph{kernel matrices} (see Table \ref{Tab_compared_kernels01}). All of these similarity measures will be used in a semi-supervised classification task and compared to 8 state-of-the-art methods on various datasets, in order to investigate if these new kernel matrices are able to accurately capture meaningful information from the graph structure.
The kernels and the semi-supervised classification technique are described in detail in the following subsections.


\begin{table}[t!]
\begin{center}
\scalebox{0.80}{%
\begin{tabular}{l l}
\hline
\textbf{Acronym} & \textbf{Method and simularity matrix}\\
\hline

BoPP & Free energy distance  \\
RSP & Randomized shortest path dissimilarity \\
SP & Shortest path distance \\
LF & Logarithmic forest distance \\
SoS & Sum-of-similarities method  \\
Q & Modularity matrix\\
LogCom & Logarithmic communicability kernel \\
Katz & Neumann kernel \\
\hline
Cov & Covariance matrix (presence on regular paths)  \\
Cor & Correlation matrix (presence on regular paths) \\
CovH & Covariance matrix (presence on hitting paths)\\
CorH & Correlation matrix (presence on hitting paths)  \\
NCov & Covariance matrix (number of occurrences on regular paths)  \\
NCor & Correlation matrix (number of occurrences on regular paths) \\
NCovH & Covariance matrix (number of occurrences on hitting paths)\\
NCorH & Correlation matrix (number of occurrences on hitting paths)  \\
\hline
\end{tabular}
}
\caption{\footnotesize{The different similarity matrices compared in this experimental section. The first eight methods are the baselines and the remaining eight are the association (covariance and correlation) measures introduced in this paper.}}
\label{Tab_compared_kernels01}
\end{center}
\end{table}

\subsection{Investigated kernels and method}

As most of the datasets used in this section are defined from their adjacency matrix $\mathbf{A} = (a_{ij})$, thus without costs assigned to edges, we define the cost matrix simply as $\mathbf{C} = (c_{ij}) = (1/a_{ij})$ \cite{Francoisse-2017}, therefore interpreting the elements of the adjacency matrix as conductances and costs as resistances. As in the standard randomized shortest paths and bag-of-paths formalism \cite{bavaud2012interpolating,FoussKernelNN-2011,Francoisse-2017,Kivimaki-2012,Mantrach-2009,Saerens-2008,Yen-08K}, the weighted adjacency matrix is defined as (see Equation (\ref{Eq_weighted_adjacency_matrix01}))
\begin{equation}
\mathbf{W} = \mathbf{P}_{\mathrm{ref}} \circ \exp[-\beta \mathbf{C}], \text{ with } \mathbf{P}_{\mathrm{ref}} = \mathbf{Diag}(\mathbf{A}\mathbf{e})^{-1} \mathbf{A}, \nonumber
\end{equation}
where $\mathbf{P}_{\mathrm{ref}}$ is the transition probabilities matrix of the natural random walk on the graph, $\mathbf{Diag}(\mathbf{v})$ is a diagonal matrix with vector $\mathbf{v}$ on its diagonal, $\mathbf{e}$ is a column vector full of $1$'s, $\circ$ is the element-wise product, and $\beta > 0$ is a hyperparameter (the inverse temperature) which will be tuned by internal cross-validation.

\subsubsection{Kernel matrices and their computation}

The 8 different kernel matrices (see Table \ref{Tab_compared_kernels01} for the definition of the acronyms) that will be compared to baseline methods are then defined as
\begin{alignat}{3}
&\mathbf{K}_\mathrm{cov} = (k^\mathrm{cov}_{ij}) &&\triangleq \left( \mathrm{cov}(\delta(i \in \wp),\delta(j \in \wp)) \right) \label{Eq_covariance_correlation_kernels01} \\
&\mathbf{K}_\mathrm{cor} = (k^\mathrm{cor}_{ij}) &&\triangleq \left( \mathrm{cor}(\delta(i \in \wp),\delta(j \in \wp)) \right) \\
&\mathbf{K}_\mathrm{covH} = (k^\mathrm{covH}_{ij}) &&\triangleq \left( \mathrm{cov}(\delta(i \in \wp^\mathrm{h}),\delta(j \in \wp^\mathrm{h})) \right) \\
&\mathbf{K}_\mathrm{corH} = (k^\mathrm{corH}_{ij}) &&\triangleq \left( \mathrm{cor}(\delta(i \in \wp^\mathrm{h}),\delta(j \in \wp^\mathrm{h})) \right) \\
&\mathbf{K}_\mathrm{Ncov} = (k^\mathrm{Ncov}_{ij}) &&\triangleq \left( \mathrm{cov}(\eta(i \in \wp),\eta(j \in \wp)) \right) \\
&\mathbf{K}_\mathrm{Ncor} = (k^\mathrm{Ncor}_{ij}) &&\triangleq \left( \mathrm{cor}(\eta(i \in \wp),\eta(j \in \wp)) \right) \\
&\mathbf{K}_\mathrm{NcovH} = (k^\mathrm{NcovH}_{ij}) &&\triangleq \left( \mathrm{cov}(\eta(i \in \wp^\mathrm{h}),\eta(j \in \wp^\mathrm{h})) \right) \\
&\mathbf{K}_\mathrm{NcorH} = (k^\mathrm{NcorH}_{ij}) &&\triangleq \left( \mathrm{cor}(\eta(i \in \wp^\mathrm{h}),\eta(j \in \wp^\mathrm{h})) \label{Eq_covariance_correlation_kernels02} \right)
\end{alignat}
The procedure for computing these kernels follows four different steps, 
\begin{enumerate}
\item First, calculate the fundamental matrix $\mathbf{Z} = ( \mathbf{I} - \mathbf{W} )^{-1} $ and choose the kernel that must be computed among (\ref{Eq_covariance_correlation_kernels01})-(\ref{Eq_covariance_correlation_kernels02}).
\item Then, depending on the kernel that is investigated, compute the needed auxiliary $z$ quantities from Results (R.2)-(R.12) or (R.15)-(R.18).
\item Depending on the kernel, compute the needed statistical moments and co-moments from Equations (\ref{Eq_cov_cor01})-(\ref{Eq_second_moment_hitting01}) or (\ref{Eq_first_moment_occurences01})-(\ref{Eq_second_moment_occurences01}).
\item Finally, compute the kernel from Equations (\ref{Eq_presence_kernels01})-(\ref{Eq_presence_kernels02}) or (\ref{Eq_occurences_kernels01})-(\ref{Eq_occurences_kernels02}).

\end{enumerate}
These eight kernels will be compared in our experiments, as detailed in the next subsections.

\subsubsection{Other investigated methods}

In this context, the introduced covariance and correlation kernels (\ref{Eq_covariance_correlation_kernels01}-\ref{Eq_covariance_correlation_kernels02}) will be compared to eight state-of-the-art methods, some of which were already investigated in previous semi-supervised and clustering tasks \cite{Francoisse-2017,ivashkin-2016,Sommer-2016,Sommer-2017,Tang-2009,Tang-2009b,Tang-2010,Zhang-2008b,Zhang-2008}:
\begin{itemize}
 \item The bag-of-paths \emph{potential distance} \cite{Francoisse-2017}, also known as the \emph{free energy distance} \cite{Kivimaki-2012}, and the \emph{randomized shortest path distance} \cite{Kivimaki-2014,Saerens-2008}, both defined in the bag-of-paths framework. These methods are respectively denoted here by \textbf{BoPP} and \textbf{RSP}, and have one hyperparameter $\beta$.
In \cite{Francoisse-2017}, it was found that the BoPP distance provided the best results overall (using almost the same methodology and datasets as those investigated in this paper). It is therefore a chalenging competitor for the 8 covariance and correlation kernels defined above.
 \item The \emph{shortest path distance} between two nodes $i$ and $j$ corresponds to the total cost along the least cost path, derived from the cost matrix $\mathbf C$. This method is denoted by \textbf{SP} and has no hyperparameter.
 \item The \emph{logarithmic forest distance} introduced in \cite{Chebotarev-2011}, is based on the matrix forest theorem \cite{Chebotarev-1997}. This method is denoted by \textbf{LF} and has one hyperparameter $\alpha$.
  \item The \emph{Neumann kernel} \cite{Scholkopf-2002}, initially proposed in \cite{Katz-1953} by Katz as a method of computing similarities, is defined as $\mathbf K=(\mathbf I-\alpha\mathbf A)^{-1}-\mathbf I$. This method is denoted by \textbf{Katz} and has one hyperparameter $\alpha$ which has to be chosen positive and smaller than the inverse of the spectral radium of $\mathbf A$, $\rho(\mathbf A)$.
  \item The \emph{logarithmic communicability kernel} \cite{ivashkin-2016} is the logarithmic version of the exponential diffusion kernel \cite{Kondor-2002}, closely related to the communicability measure \cite{Estrada-2008}, and defined as $\mathbf K=\mathrm{ln}(\mathrm{expm}{(\alpha\mathbf A)})$, where $\mathrm{expm}$ is the matrix exponential. This method is denoted by \textbf{LogCom} and has one positive hyperparameter $\alpha$.
\item The \emph{modularity matrix} $\mathbf{Q}$ \cite{Newman-2006,Newman-2018} defined as $\mathbf Q = \mathbf A - \frac{\mathbf{dd}^{\text T}}{\mathrm{vol}}$ where $\mathbf{d}$ contains the node degrees and $\mathrm{vol}$ is the volume of the graph. This method is denoted by \textbf{Q} and has no hyperparameter.
\item The \emph{sum-of-similarities} method \cite{FoussKernelNN-2011,mantrach2011semi}, which is based on the \emph{regularized commute time kernel}. This method differs from the previous ones, as it uses a simple label propagation technique in order to classify nodes. It is both computationally efficient and competitive in terms of accuracy \cite{FoussKernelNN-2011,mantrach2011semi,Francoisse-2017}. This method is denoted by \textbf{SoS} and has one hyperparameter $\alpha$.
\end{itemize}

\subsubsection{The semi-supervised classification task}

The methodology for comparing the different measures on semi-supervised tasks closely follows\footnote{We thank the authors of this paper for providing the code and the datasets.} \cite{Francoisse-2017}. Therefore the procedure is summarized; for a more complete description of the methods, see reference \cite{Francoisse-2017}, Section 7.

The classification method consists in extracting 5 graph feature vectors\footnote{We arbitrarily extracted 5 dimensions but performed experiments with more dimensions with similar conclusions.} from these kernel and similarity/dissimilarity matrices by classical multidimensional scaling in order to use them as input to a linear support vector machine (SVM) for classification. Thus, information is extracted from the structure of the graph in an unsupervised way.

More precisely, first, following classical multidimensional scaling \cite{Borg-1997,Cox-2001}, each dissimilarity matrix $\mathbf{D}$ is transformed into a kernel matrix by centering, $\mathbf{K} = -\tfrac{1}{2} \mathbf{H} \mathbf{D}^{(2)} \mathbf{H}$ where $\mathbf{H} = \mathbf{I} - \mathbf{e} \mathbf{e}^{\text T} / n$ is the centering matrix, $\mathbf{e}$ is a column vector full of 1's and $\mathbf{D}^{(2)}$ is the matrix of (elementwise) squared distances \cite{Borg-1997,Cox-2001}.
Then, the spectral decomposition of each kernel and similarity matrix is computed and the $p$ dominant eigenvectors $\mathbf{u}_{k}$ (column vectors) are sorted by decreasing eigenvalue $\lambda_{1} \ge \lambda_{2} \ge \cdots \ge \lambda_{p}$. Eigenvectors corresponding to negative eigenvalues are then eliminated. At this stage, we tested two different options for extracting the node features.
\begin{itemize}
    \item Either the dominant eigenvectors $\mathbf{u}_{k}$ are weighted by the square root of their corresponding eigenvalues, $\sqrt{\lambda_{k}}$, and concatenated in order to build the data matrix $\mathbf{X}$ containing node features on its rows, $\mathbf{X} = [ \sqrt{\lambda_{1}} \mathbf{u}_{1}, \sqrt{\lambda_{2}} \mathbf{u}_{2}, \dots  ]$, to be injected as input to a SVM. This is exactly multidimentional scaling limited to $p$ dimensions.
    \item Or the eigenvectors are simply concatenated $\mathbf{X} = [ \mathbf{u}_{1}, \mathbf{u}_{2}, \dots  ]$ and then each row of $\mathbf{X}$ is normalized, $\mathbf{x}_{i} \leftarrow \mathbf{x}_{i} / \| \mathbf{x}_{i} \|_{2}$, so that each node feature vector is of unit length. This corresponds to the projection of the feature vector on the sphere of radius 1 centered at the origin. This way of defining the node feature vectors removes the effect of the size of the vector, which works better when the angle between the node vectors is more relevant than their distance, like, for instance, for covariance measures. 
\end{itemize}
For simplicity, we only report the results of the \emph{best option} for each method, according to the Nemenyi test described later.

This setting is inspired by the work of  Zhang et al. \cite{Zhang-2008b,Zhang-2008} as well as Tang et al. \cite{Tang-2009,Tang-2009b,Tang-2010} who compute the dominant eigenvectors (a ``latent space") of graph kernels or similarity matrices and then input them into a supervised classification method, such as a logistic regression or a SVM, to categorize the unlabeled nodes. Notice that these techniques based on similarities and eigenvectors extraction sometimes allow to scale to large graphs, depending on the kernel \cite{Devooght-2014}.

\subsection{Experimental settings}
\label{exp_set}
As already mentioned, the experimental methodology is inspired by \cite{Francoisse-2017} and briefly summarized here; see this reference for further details.

\subsubsection{datasets}

The different classification methods will be compared on 14 well-known network datasets, already used in previous experimental comparisons.

\begin{itemize}
\item {\bf WebKB (4 datasets)}. These datasets \cite{macskassy2007classification} come from networks of co-citation between webpages of computer science departments of 4 different universities: {\bf webKB-texas}, {\bf webKB-washington}, {\bf webKB-wisconsin} and {\bf webKB-cornell}. Each page of these website has been labeled manually to form six different classes: course, department, faculty, project, staff and student.

\item {\bf 20 Newsgroups (9 subsets)}. The \emph{Newsgroup} dataset consists of 20.000 documents taken from 20 discussion groups of the Usenet diffusion list \cite{lichman2013uci}. Nine subsets have been extracted from this data (for details, see \cite{Yen-07,yen2009graph}): {\bf news-2cl-1}, {\bf news-2cl-2}, {\bf news-2cl-3}, {\bf news-3cl-1}, {\bf news-3cl-2}, {\bf news-3cl-3}, {\bf news-5cl-1}, {\bf news-5cl-2}, and {\bf news-5cl-3}. As their names suggest, there are different numbers of classes/topics in these datasets (e.g., {\bf news-3cl-1} contains documents from 3 different topics considered as classes). For each of these datasets, a sparse graph structure was derived from the \emph{term-document} matrix $\mathbf{T} \triangleq (t_{ij})$, with $t_{ij}$ being the \emph{tf-idf} score \cite{manning2008introduction} of term $i$ in document $j$. The corresponding adjacency matrix is then obtained by $\mathbf{A}= \mathbf{T}^\top \mathbf{T}$.

\item {\bf IMDB}. This dataset comes from the well-known \emph{Internet Movie Database} \cite{macskassy2007classification}. Nodes represent movies and the adjacency matrix contains the number of production companies two movies have in common. There are only two labels in this dataset: box-office hit or not.

\end{itemize}

\subsubsection{Comparisons on semi-supervised classification}

The different classification methods are compared in terms of classification accuracy on semi-supervised tasks where a subset of nodes of the graph is kept unlabeled (their label is hidden to the classifier). The classification model then predicts the label of these unlabeled nodes and its prediction is compared to the true label which was hidden.

In order to reduce variance in accuracy, methods are tested by using a standard $5 \times 5$ nested cross-validation methodology. Each external cross-validation contains 5 folds, and methods are tested with a labelling rate of $20\%$ ($80\%$ of the labels are hidden and then predicted by the classifier trained on the labeled $20\%$). To tune hyperparameters, an internal 5-fold cross-validation on the training fold is performed, by taking $4/5$ of the training fold as labelled and $1/5$ as unlabelled. The average accuracy obtained from the external cross-validation folds is then computed for each method. The whole cross-validation procedure is repeated 5 times for different random permutations of the data, inducing different sets of labeled/unlabeled nodes. The \emph{final accuracy} of the classifier on the investigated dataset is then obtained by averaging the results over the five repetitions, and is reported in Table \ref{table:classres}. 

Concerning the hyperparameters, the bag-of-paths-type methods and logarithmic forest distance (LF) investigate tuning values of $\{10^{-6},10^{-5},10^{-4},10^{-3}, \\ 10^{-2},10^{-1},1,10\}$, the Katz kernel and the sum-of-similarities (SoS) method $\{0.1,0.2,0.3,0.4,0.5,0.6,0.7,0.8,0.9\}$ and the logarithmic communicability kernel (LogCom) $\{0.01, 0.1, 0.5, 1, 5, 10\}$. For the SVM, the margin hyperparameter is tuned on the set of values $\{10^{-2},10^{-1}, 1, 10, 100\}$. External and internal cross-validation folds are kept identical for all methods and all repetitions.

\subsection{Results and discussion}
\begin{table}[t!]
\footnotesize
\begin{center}
\scalebox{0.8}{
\begin{tabular}{lcccccccc}
\hline
\textbf{Classif. method}: & BoPP & RSP & SP & LF & SoS & Q & LogCom  &Katz \\
\textbf{Dataset}:&   &   &   &   &  &  &   &    \\
\hline
webKB-texas & 73.35 &75.82 &64.75 &75.28 &73.26 &73.01 &75.45 &64.98 \\
webKB-washington & 68.15 &70.35 &56.47 &\textbf{71.19} &61.72 &62.52 &69.92 &67.25\\
webKB-wisconsin &74.33 &74.02 &63.99 &74.45 &73.88 &73.42 &74.99 &73.46 \\
webKB-cornell &57.20 &57.31 &48.76 &58.24 &57.43 &50.71 &\textbf{59.16} &52.88  \\
imdb &75.31 &76.68 &75.22 &76.41 &\textbf{78.12} &74.37 &76.28 &73.75  \\
news-2cl-1 &96.79 &96.14 &93.46 &96.83 &91.09 &95.85 &96.18&95.84 \\
news-2cl-2 &91.21 &90.16 &90.15 &89.62 &87.70 &91.22 &91.18 &91.49 \\
news-2cl-3 &95.98 &95.66 &95.90 &95.14 &94.20 &95.78 &95.43 &95.50 \\
news-3cl-1 &92.53 &93.08 &93.18 &93.00 &88.93 &93.02 &94.00 &93.56 \\
news-3cl-2 &93.45 &92.56 &89.53 &91.33 &88.01 &92.63 &92.17 &93.30  \\
news-3cl-3 &\textbf{93.66} &93.06 &91.71 &91.19 &88.29 &91.20 &91.49 &90.37 \\
news-5cl-1 &\textbf{88.70} &87.81 &86.34 &86.63 &84.84 &77.04 &86.30&79.49  \\
news-5cl-2 &\textbf{82.32} &81.81 &78.42 &80.49 &78.80 &75.97 &79.04&69.25 \\
news-5cl-3 &80.27 &\textbf{80.30} &73.11 &79.45 &79.22 &76.51 &78.65&68.65 \\
\hline \hline
\textbf{Classif. method}: & Cov & Cor & CovH & CorH & NCov & NCor & NCovH & NCorH \\
\textbf{Dataset}:&   &   &   &   &  &  &   &   \\
\hline
webKB-texas & 75.52 &75.76 &76.57 &76.48 &75.63 &75.54 &\textbf{76.90} &76.65  \\
webKB-washington &69.47 &66.59 &67.21 &67.30 &68.09 &68.57 &66.44 &66.62 \\
webKB-wisconsin  &74.25 &\textbf{75.56} &74.37 &74.07 &73.21 &73.13 &74.14 &73.85 \\
webKB-cornell & 55.75 &58.83 &58.86 &58.13 &52.90 &53.32 &58.02 &58.97 \\
imdb &76.04 &77.57 &77.00 &76.57 &76.42 &76.39 &76.67 &76.67 \\
news-2cl-1  &96.61 &96.73 &96.83 &96.53 &96.05 &96.05 &96.80 &\textbf{97.26}\\
news-2cl-2 &91.42 &\textbf{91.86} &91.07 &90.84 &91.66 &91.83 &90.67 &91.17 \\
news-2cl-3 &96.49 &96.45 &96.45 &96.30 &96.53 &\textbf{96.75} &96.42 &96.55 \\
news-3cl-1 &93.72 &93.74 &93.76 &93.68 &93.43 &\textbf{94.06} &93.80 &93.60 \\
news-3cl-2 &93.72 &93.39 &93.11 &93.04 &\textbf{93.79} &93.78 &92.98 &92.87 \\
news-3cl-3 &91.69 &91.45 &91.51 &91.81 &91.90 &91.85 &91.76 &91.92 \\
news-5cl-1 &83.28 &83.43 &86.38 &86.45 &82.91 &82.48 &86.33 &86.18 \\
news-5cl-2  &75.03 &75.06 &77.07 &77.56 &73.18 &75.02 &77.25 &77.50 \\
news-5cl-3  &78.39 &78.35 &76.56 &76.56 &76.86 &76.85 &76.26 &76.22 \\
\hline
\end{tabular}}
\end{center}
\caption{\footnotesize{Classification accuracy in percent for the various classification methods obtained on the different datasets. For each dataset and method, the final accuracy is obtained by averaging over 5 repetitions of a standard cross-validation procedure. Each repetition consists of a nested cross-validation with 5 external folds (test sets, for validation) on which the accuracy of the classifier is averaged, and 5 internal folds (for parameter tuning). The best performing method is highlighted in boldface for each dataset.}}
\label{table:classres}
\end{table}
\begin{table}[t!]
\footnotesize
\begin{center}
\begin{tabular}{lcr}
\hline
\textbf{Method} & \textbf{Position} & \textbf{Score} \\
\hline
CovH & 1    & 152  \\
Cor & 2    & 149   \\
BoPP & 3 & 147 \\
NCorH & 4 & 144 \\
RSP & 5 & 143\\
LogCom & 6 & 137\\
CorH & 7 & 136\\
NCovH & 7 & 136\\
Cov & 8 & 131\\
NCor & 8 & 131\\
LF & 9 & 128\\
NCov & 10 & 121\\
SoS & 11 & 76\\
Katz & 12 & 64\\
Q & 13 & 61\\
SP & 14 & 60\\
\hline
\end{tabular}
\end{center}
\caption{\footnotesize{Overall position of the different classification techniques (see Table \ref{Tab_compared_kernels01}) according to Borda$'$s method performed across all datasets (the higher the score, the better).}}
\label{table:Borda}
\end{table}

As already mentioned, Table \ref{table:classres} reports average classification accuracy in percent for all the investigated methods on the different datasets, for labeling rates of 20\%. The method performing best is highlighted in boldface for each dataset. Moreover, a simple Borda ranking of the methods is performed and shown in Table \ref{table:Borda}. This ranking provides a score to each method equal to the sum of its ranks over all the datasets, where the methods are sorted in ascending order of classification accuracy. Thus, the higher the rank, the higher the accuracy on the dataset. The best method overall in this context is the one showing the highest Borda score. 

As can be seen in Table \ref{table:Borda}, the covariance matrix based on the presence of nodes on hitting paths (CovH), the correlation matrix based on the presence of nodes on regular paths (Cor) and the bag-of-paths free energy potential distance (BoPP) obtain the best results overall. Besides, the randomized shortest path dissimilarity (RSP), the logarithmic communicability kernel (LogCom) and the other covariance and correlation measures, except the covariance matrix based on the number of occurrences on regular paths (NCov), provide good results superior to those obtained by the logarithmic forest distance (LF) baseline. We can also observe that the sum-of-similarities (SoS), the Neumann kernel (Katz), the modularity matrix (Q) and the shortest path distance (SP) obtain much worse results in comparison with the other methods. However, when examining the raw results of Table \ref{table:classres}, it can be seen that the best method is dataset-dependent and that no obvious pattern is present. 

Consequently, in order to rate globally the results of each method, a nonparametric Friedman-Nemenyi statistical test \cite{Demsar2006} allowing to make comparisons across all the datasets is investigated. At first, we run a Friedman test \cite{Friedman:1937,Friedman:1990} on our results. This will tell us whether at least one classification method is significantly different from the others. We obtain a $p$-value of $5 \, 10^{-6}$. This $p$-value is lower than the threshold $\alpha$ of $0.05$ and we can reject the null hypothesis, at least at this level. 

\begin{figure}[t!]
\begin{center}
       \includegraphics[width = 0.7\textwidth, trim=0 20 0 0, clip=true]{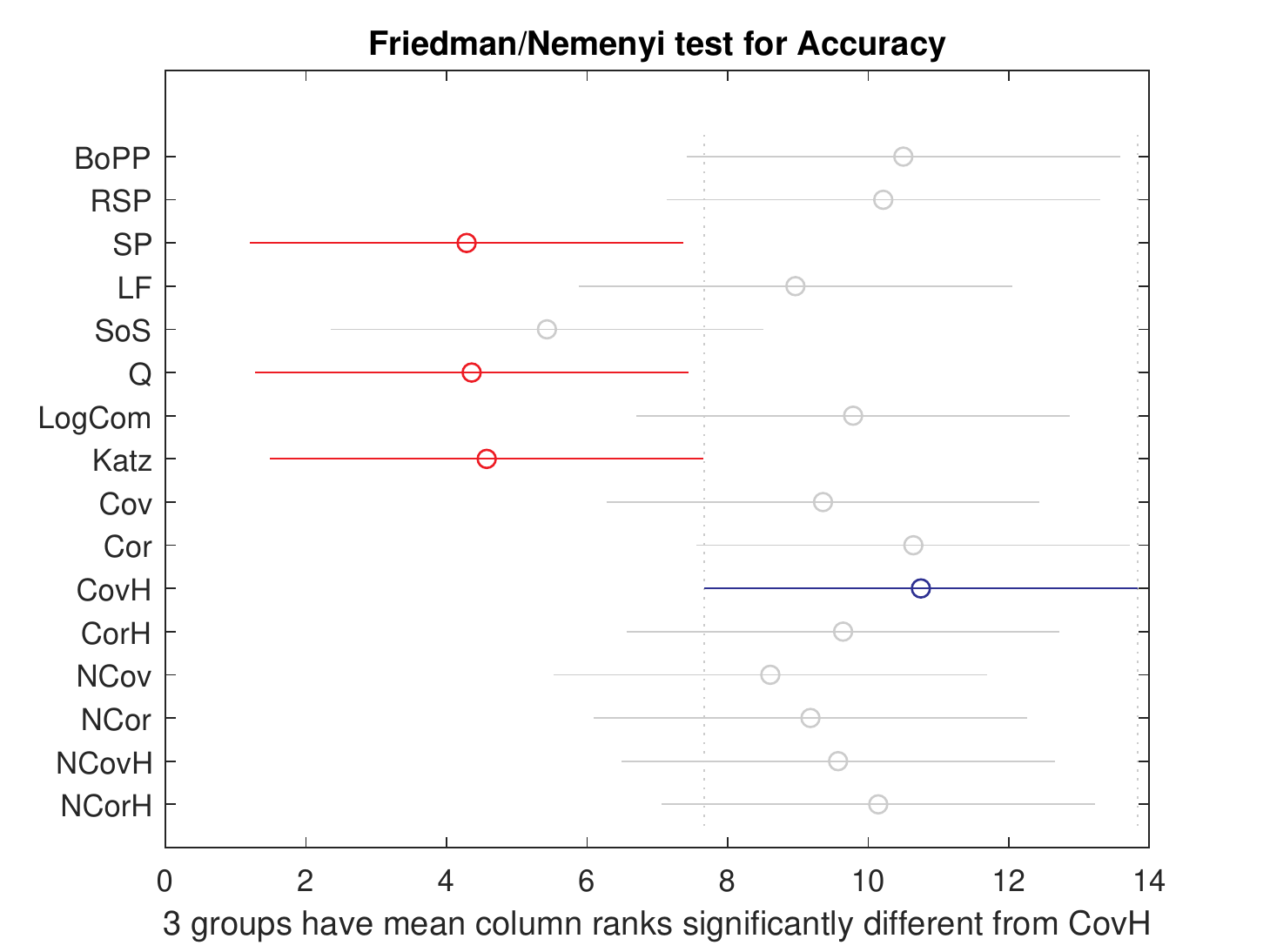}
\end{center}
\caption{\footnotesize{Mean ranks and 95\% Nemenyi confidence intervals for the 16 methods (see Table \ref{Tab_compared_kernels01}) across the 14 datasets. Two methods are considered as significantly different if their confidence intervals do not overlap. The higher the rank, the best the method. The worse method (SP) and the best method (CovH) overall are highlighted.}}
\label{fig:meanRank}
\end{figure}

Then, we perform a multiple comparison with the Nemenyi test \cite{Demsar2006,Nemenyi:1963}. The results of this test are illustrated in Figure \ref{fig:meanRank} and are similar to those provided by the Borda ranking. The figure confirms that the CovH, the Cor and the BoPP provide good results, which are significantly superior to the results obtained by the modularity matrix Q and the shortest path distance SP. As concerns the other covariance and correlation measures, we cannot say that they are significantly different from the other baselines. This is partly because the Friedman-Nemenyi test is rather conservative, especially when comparing many different models.

Therefore, to obtain more precise information concerning the relative performance of the methods, we decided to also perform a Wilcoxon signed-ranks tests \cite{Demsar2006,Wilcoxon:1945} for matched data with a threshold $\alpha$ of $0.05$. 

These paired tests show that all the introduced covariance and correlation methods, the BoPP, the RSP, the LF and the LogCom are significantly better than the modularity matrix Q, the shortest path distance SP and the Neumann kernel (Katz). The tests also show that all these methods except the Cov, the NCov and the NCor obtain significantly better results than the SoS.


In summary, the experiments showed that the majority of the introduced covariance and correlation measures achieve good results in comparison with our eight baselines on the investigated datasets.
However, we cannot conclude that the introduced methods perform better than the other baselines (BoPP, RSP, LF and LogCom): only some of these methods (CovH and Cor) are globally better ranked than them. Nevertheless, we can state that they perform at least as well as these baselines on the investigated datasets, which is already an excellent result as some of the baselines performed very well in previous experimental comparisons, in both semi-supervised classification and clustering tasks \cite{Francoisse-2017,ivashkin-2016,Sommer-2016,Sommer-2017,Tang-2009,Tang-2009b,Tang-2010,Zhang-2008b,Zhang-2008} .

\section{Conclusion}
\label{Sec_conclusion}

This paper derived a series of useful mathematical expressions for computing the expectation of (co-)presence and of the number of (co-)occurrences of nodes on paths sampled from a network. These quantities can then be used for defining similarity measures between nodes as well as betweenness centrality measures. Most of the derived similarity measures are positive semi-definite and can therefore be considered as kernels on a graph and used with kernel-based methods for solving pattern recognition tasks such as node clustering and supervised classification. The main intuition behind these measures is that two nodes are considered as closely related if they often co-occur on the same (preferably short) paths in the network.
Experiments on semi-supervised classification tasks have shown that the introduced quantities provide competitive results on the investigated datasets, within a clear theoretical framework.

Note that, as in \cite{Fouss-2016,Lebichot-2014}, the different methods could easily be extended in order to provide similarity and centrality measures between groups of nodes, which is left for further work.

Another application that is left for future work is the task of embedding graphs in low-dimensional spaces \cite{Cai-2018,Zhang-2018}. Indeed, such a representation can easily be deduced from our measures by applying, e.g., classical multidimensional scaling to the defined kernels. Interestingly, the recently introduced DeepWalk technique \cite{Perozzi-2014} (an example of representation learning applied to graphs) and variants seem to provide accurate low-dimensional representations of the nodes of the network. Quoting \cite{Perozzi-2014}, ``DeepWalk uses local information obtained from truncated random walks to learn latent representations by treating walks as the equivalent of sentences" (in natural language processing). This is similar in spirit to the techniques studied in this paper with an important difference: our measures are computed in closed form through matrix operations whereas DeepWalk uses sampled paths and a neural network to build the representation. Therefore, we plan to work on a thorough experimental comparison between representation learning based techniques and our proposed methods, on semi-supervised classification and graph embedding tasks.

However, the main drawback of the introduced techniques is the fact that they require the inversion of a $n \times n$ matrix so that they do not scale well on large graphs. Thus, another interesting further work would be the study of co-occurrence measures, defined this time on truncated paths, such as in \cite{mantrach2011semi,Sarkar2007}. This should allow to scale to medium to large, sparse, graphs.


\section*{Acknowledgements}

\noindent This work was partially supported by the Immediate and the Brufence projects funded by InnovIris (Brussels Region), as well as former projects funded by the Walloon region, Belgium. We thank these institutions for giving us the opportunity to conduct both fundamental and applied research.

\begin{center}
\rule{2.5in}{0.01in} 
\par\end{center}

\section*{Appendix: proofs of the main results}
\appendix
\numberwithin{equation}{section}

\section{Proofs of results of Section \ref{section_weight_formulae}}
\label{ App_A}

This appendix contains the proofs of the results stated in Section \ref{section_weight_formulae}. 

\subsection{Results involving one intermediary node}

\paragraph{(R.1).} Let $\mathcal{P}_{st}(\ell)$ be the set of all regular paths from $s$ to $t$ of length exactly equal to $\ell$. We easily obtain that
\begin{equation}
    \sum_{\wp^{\ell}_{st} \in \mathcal{P}_{st}(\ell)} w(\wp^{\ell}_{st})
    = \sum_{i_{1}, i_{2}, \dots, i_{\ell - 1} = 1}^{n} w_{s i_{1}} w_{i_{1} i_{2}} \cdots w_{i_{\ell - 1} t}
    = [\mathbf{W}^\ell]_{st}.
\end{equation}
Therefore, from (\ref{eq:fundamentalMatrixDefinition01}),
\begin{align}
w(\mathcal{P}_{st}) &= \sum_{\wp_{st} \in \mathcal{P}_{st}} w(\wp_{st}) =  \sum_{\ell = 0}^\infty \sum_{\wp^{\ell}_{st} \in \mathcal{P}_{st}(\ell)} w(\wp^{\ell}_{st}) \nonumber \\
&= \sum_{\tau = 0}^\infty [\mathbf{W}^\tau]_{st} = \left[\sum_{\tau = 0}^\infty \mathbf{W}^\tau \right]_{st} = \left[ \mathbf{Z} \right]_{st} = z_{st} \tag{R.1}
\end{align}

\paragraph{(R.2).} Observe that there is a bijection between  $\mathcal{P}_{st}$ and $\mathcal{P}^\mathrm{h}_{st} \circ \mathcal{P}_{tt}$. In other words, any regular path from $s$ to $t$ can be decomposed uniquely into a hitting sub-path from $s$ to $t$, where node $t$ is reached for the first time, followed by a regular sub-path (a cycle) connecting $t$ to itself. We then have $w(\wp_{st}) = w(\wp^\mathrm{h}_{st}) w(\wp_{tt})$ for the weight of corresponding paths. This implies that
\begin{align}
&w(\mathcal{P}^\mathrm{h}_{st})  w(\mathcal{P}_{tt}) = \Bigg( \sum_{\wp^\mathrm{h}_{st} \in \mathcal{P}^\mathrm{h}_{st}} w(\wp^\mathrm{h}_{st}) \Bigg) \Bigg( \sum_{\wp_{tt} \in \mathcal{P}_{tt}} w(\wp_{tt}) \Bigg) \nonumber \\
&= \sum_{\wp^\mathrm{h}_{st} \in \mathcal{P}^\mathrm{h}_{st}} \sum_{\wp_{tt} \in \mathcal{P}_{tt}} w(\wp^\mathrm{h}_{st}) w(\wp_{tt})
= \sum_{\wp_{st} \in \mathcal{P}_{st}} w(\wp_{st}) = w(\mathcal{P}_{st}), \nonumber
\end{align}
and, from (R.1), we get the result (R.2), i.e., $z^\mathrm{h}_{st} = z_{st}/z_{tt}$.

\paragraph{(R.3).} Similarly to (R.2), there exists a bijection between $\mathcal{P}^\mathrm{h}_{si} \circ \mathcal{P}_{it}$ and $\mathcal{P}^{(+i)}_{st}$. Indeed, any path from $s$ to $t$ visiting $i$ can be decomposed uniquely into a hitting sub-path from $s$ to $i$, where node $i$ is visited for the first time, followed by a regular sub-path from $i$ to $t$. Consequently, we have $w(\wp_{st}) = w(\wp^\mathrm{h}_{si}) w(\wp_{it})$ for corresponding paths. Thus,
\begin{align}
w\left(\mathcal{P}^{(+i)}_{st}\right) &= \sum_{\wp_{st} \in \mathcal{P}_{st}} \delta(i \in \wp_{st}) w(\wp_{st}) \nonumber \\
&= \sum_{\wp^\mathrm{h}_{si} \in \mathcal{P}^\mathrm{h}_{si}} \sum_{\wp_{it} \in \mathcal{P}_{it}} w(\wp^\mathrm{h}_{si}) w(\wp_{it}) = z^\mathrm{h}_{si} z_{it}. \tag{R.3}
\end{align}

\paragraph{(R.4).} From (R.3), we observe that 
\begin{align}
w\left(\mathcal{P}^{(-i)}_{st}\right) &= \sum_{\wp_{st} \in \mathcal{P}_{st}} \delta(i \notin \wp_{st}) w(\wp_{st}) \nonumber \\
&= \sum_{\wp_{st} \in \mathcal{P}_{st}} (1 - \delta(i \in \wp_{st})) w(\wp_{st}) = z_{st} - z^\mathrm{h}_{si} z_{it}. \tag{R.4}
\end{align}
and the result is also valid for $s = t$.

\paragraph{(R.5).} Let us now consider results involving hitting paths. The derivation is similar to the proof of (R.4) and (R.2). Considering $i\neq t$,
\begin{align*}
w\left(\mathcal{P}^{(-i)}_{st}\right)  &= \sum_{\wp_{st} \in \mathcal{P}_{st}} \delta(i \notin \wp_{st}) w(\wp_{st}) \\
&=\sum_{\wp^\mathrm{h}_{st} \in \mathcal{P}^\mathrm{h}_{st}}  \sum_{\wp_{tt} \in \mathcal{P}_{tt}} \delta(i \notin \wp^\mathrm{h}_{st}) w(\wp^\mathrm{h}_{st}) \, \delta(i \notin \wp_{tt}) w(\wp_{tt}) \\
&= w\left(\mathcal{P}^{\mathrm{h}(-i)}_{st}\right)  w\left(\mathcal{P}^{(-i)}_{tt}\right).
\end{align*}

Therefore, by isolating $w\left(\mathcal{P}^{\mathrm{h}(-i)}_{st}\right)$ and using (R.4) as well as (R.2),
\begin{equation}
w\left(\mathcal{P}^{\mathrm{h}(-i)}_{st}\right)  = \frac{z^{(-i)}_{st}}{z^{(-i)}_{tt}} =   \frac{z_{st} - z^\mathrm{h}_{si} z_{it}}{z_{tt} - z^\mathrm{h}_{ti} z_{it}} =  \frac{z^\mathrm{h}_{st} - z^\mathrm{h}_{si} z^\mathrm{h}_{it}}{1 - z^\mathrm{h}_{ti} z^\mathrm{h}_{it}}, \tag{R.5}
\end{equation}
where we divided the numerator and the denominator by $z_{tt}$.
Now, if $i = t$, we trivially have that $ \delta(i \notin \wp^\mathrm{h}_{st}) = 0$ and therefore $z^{\mathrm{h}(-t)}_{st} = 0$.

\paragraph{(R.6).} Still for hitting paths, if $i\neq t$, we obtain from $w\left(\mathcal{P}^{\mathrm{h}}_{st}\right) = w\left(\mathcal{P}^{\mathrm{h}(+i)}_{st}\right) + w\left(\mathcal{P}^{\mathrm{h}(-i)}_{st}\right)$ and from the previous result (R.5),
\begin{align*}
w\left(\mathcal{P}^{\mathrm{h}(+i)}_{st}\right) &= w\left(\mathcal{P}^{\mathrm{h}}_{st}\right) - w\left(\mathcal{P}^{\mathrm{h}(-i)}_{st}\right) = z^{\mathrm{h}}_{st} - z^{\mathrm{h}(-i)}_{st} \\
&= \frac{(1 - z^\mathrm{h}_{ti} z^\mathrm{h}_{it})z^\mathrm{h}_{st}}{1 - z^\mathrm{h}_{ti} z^\mathrm{h}_{it}}  - \frac{z^\mathrm{h}_{st} - z^\mathrm{h}_{si} z^\mathrm{h}_{it}}{1 - z^\mathrm{h}_{ti} z^\mathrm{h}_{it}} \\
&=   \frac{(z^\mathrm{h}_{si} - z^\mathrm{h}_{st} z^\mathrm{h}_{ti})z^\mathrm{h}_{it}}{1 - z^\mathrm{h}_{ti} z^\mathrm{h}_{it}}= z^{\mathrm{h}(-t)}_{si} z^\mathrm{h}_{it}. \tag{R.6}
\end{align*}
Moreover, if $i = t$, obviously $z^{\mathrm{h}(+i)}_{st} = z^\mathrm{h}_{st}$.

\subsection{Results involving two intermediary nodes}

\paragraph{(R.7).} Assuming $i \neq j$, let $\mathcal{P}^{(i<j)}_{st}$ be the set of all regular paths containing $i$ and $j$ with node $i$ appearing first on the path, that is, before the first visit to node $j$. Then
\begin{align*}
w\left(\mathcal{P}^{(+\{i,j\})}_{st}\right) = w\left(\mathcal{P}^{(i<j)}_{st}\right) + w\left(\mathcal{P}^{(j<i)}_{st}\right)
\end{align*}
We can observe that there exists a bijection between $\mathcal{P}^{(i<j)}_{st}$ and  $(\mathcal{P}^{\mathrm{h}(-j)}_{si} \circ \mathcal{P}^{\mathrm{h}}_{ij} \circ \mathcal{P}_{jt})$ with corresponding weights. Indeed, each path $\wp$ from $s$ to $t$, visiting node $i$ first and then node $j$, can be decomposed into a first hitting sub-path from $s$ to the first occurrence of node $i$ on $\wp$, followed by a second hitting sub-path from $i$ to the first occurrence of node $j$, and finally a third regular sub-path from $j$ to $t$. Thus,
\begin{align*}
w\left(\mathcal{P}^{(i<j)}_{st}\right) &= w\left(\mathcal{P}^{\mathrm{h}(-j)}_{si}\right) w\left(\mathcal{P}^{\mathrm{h}}_{ij}\right)  w\left(\mathcal{P}_{jt} \right) = z^{\mathrm{h}(-j)}_{si} z^\mathrm{h}_{ij} z_{jt}.
\end{align*}
We therefore obtain from (R.6), for $i \neq j$,
\begin{align*}
w\left(\mathcal{P}^{(+\{i,j\})}_{st}\right) = z^{\mathrm{h}(-j)}_{si} z^\mathrm{h}_{ij} z_{jt} + z^{\mathrm{h}(-i)}_{sj} z^\mathrm{h}_{ji} z_{it} = z^{\mathrm{h}(+i)}_{sj} z_{jt} + z^{\mathrm{h}(+j)}_{si} z_{it}, \tag{R.7}
\end{align*}
which is the desired result.

\paragraph{(R.8).} We will derive three different expressions for computing the same quantity. For the first expression, by assuming $i \neq j$ and using (R.1), (R.3) and (R.7), we obtain
\begin{align}
w\left(\mathcal{P}^{(-\{i,j\})}_{st}\right) &= \sum_{\wp_{st} \in \mathcal{P}_{st}} (1-\delta(i \in \wp_{st})) (1- \delta(j \in \wp_{st})) \, w(\wp_{st}) \notag \\
&= z_{st} -  z^\mathrm{h}_{si} z_{it} -  z^\mathrm{h}_{sj} z_{jt} + w\left(\mathcal{P}^{(+\{i,j\})}_{st}\right) \notag \\
&= z_{st} -  z^\mathrm{h}_{si} z_{it} -  z^\mathrm{h}_{sj} z_{jt} + (z^{\mathrm{h}(+i)}_{sj} z_{jt} + z^{\mathrm{h}(+j)}_{si} z_{it}) \notag \\
&= z_{st} -  (z^\mathrm{h}_{si} - z^{\mathrm{h}(+j)}_{si}) z_{it} -  (z^\mathrm{h}_{sj} - z^{\mathrm{h}(+i)}_{sj}) z_{jt} \label{Eq_R8_intermediate01} \\
&= z_{st} - z^{\mathrm{h}(-j)}_{si} z_{it}  - z^{\mathrm{h}(-i)}_{sj} z_{jt}.  \tag{R.8.1}
\end{align}
Note that expression (\ref{Eq_R8_intermediate01}) is obtained by rearranging the terms. Moreover, the last expression (R.8.1) is either obtained by direct calculation, or by observing that $z^{\mathrm{h}}_{si} = z^{\mathrm{h}(+j)}_{si} + z^{\mathrm{h}(-j)}_{si}$.

Let us derive still another expression for this quantity. As a regular path from $s$ to $t$ visiting node $j$ can be decomposed uniquely into a hitting path from $s$ to $j$ (visiting $j$ for the first time) followed by a regular path from $j$ to $t$, we have
\begin{align}
w\left(\mathcal{P}^{(-\{i,j\})}_{st}\right) &= \sum_{\wp_{st} \in \mathcal{P}_{st}} (1-\delta(i \in \wp_{st})) (1- \delta(j \in \wp_{st})) \, w(\wp_{st}) \notag \\
&= \sum_{\wp_{st} \in \mathcal{P}_{st}} (1-\delta(i \in \wp_{st})) \, w(\wp_{st}) \notag \\
& \quad - \sum_{\wp_{st} \in \mathcal{P}_{st}} (1-\delta(i \in \wp_{st}))  \delta(j \in \wp_{st}) \, w(\wp_{st}) \notag \\
&= w\left(\mathcal{P}^{(-i)}_{st}\right) - \sum_{\wp_{st} \in \mathcal{P}_{st}} \delta(i \notin \wp_{st})  \delta(j \in \wp_{st}) \, w(\wp_{st}) \notag \\
&= w\left(\mathcal{P}^{(-i)}_{st}\right) - \sum_{\wp^\mathrm{h}_{sj} \in \mathcal{P}^\mathrm{h}_{sj}}  \sum_{\wp_{jt} \in \mathcal{P}_{jt}} \delta(i \notin \wp^\mathrm{h}_{sj}) w(\wp^\mathrm{h}_{sj}) \, \delta(i \notin \wp_{jt}) w(\wp_{jt}) \notag \\
&= w\left(\mathcal{P}^{(-i)}_{st}\right) - w\left(\mathcal{P}^{\mathrm{h}(-i)}_{sj}\right)  w\left(\mathcal{P}^{(-i)}_{jt}\right) \notag \\
&= z^{(-i)}_{st} - z^{\mathrm{h}(-i)}_{sj} z^{(-i)}_{jt}.  \tag{R.8.2}
\end{align}

It can be shown in the same way that, symmetrically,
\begin{equation}
w\left(\mathcal{P}^{(-\{i,j\})}_{st}\right) = z^{(-j)}_{st} - z^{\mathrm{h}(-j)}_{si} z^{(-j)}_{it}.  \tag{R.8.3}
\end{equation}

\paragraph{(R.9).} Let us consider hitting paths now. It is obvious that $w\left(\mathcal{P}^{\mathrm{h}(-\{i,j\})}_{st} \right) = 0$ if $i = t$ or $j = t$. Therefore, let us consider $t \neq i \neq j \neq t$. Similarly to (R.2), we decompose paths $\wp_{st}$ in $\wp_{st}^{\mathrm{h}} \circ \wp_{tt}$,
\begin{align*}
w\left(\mathcal{P}^{(-\{i,j\})}_{st} \right) &= \sum_{\wp_{st} \in \mathcal{P}_{st}} \delta(i \notin \wp_{st}) \delta(j \notin \wp_{st}) \, w(\wp_{st}) \\ 
&= \sum_{\wp^\mathrm{h}_{st} \in \mathcal{P}^\mathrm{h}_{st}} \delta(i \notin \wp^\mathrm{h}_{st}) \delta(j \notin \wp^\mathrm{h}_{st}) \, w(\wp^\mathrm{h}_{st})  \sum_{\wp_{tt} \in \mathcal{P}_{tt}} \delta(i \notin \wp_{tt}) \delta(j \notin \wp_{tt}) \, w(\wp_{tt}) \\
&= z^{\mathrm{h}(-\{i,j\})}_{st} z^{(-\{i,j\})}_{tt}.
\end{align*}
Consequently, $z^{\mathrm{h}(-\{i,j\})}_{st} =  z^{(-\{i,j\})}_{st}/z^{(-\{i,j\})}_{tt}$ which proves (R.9); thus, with the help of (R.8.1)-(R.8.3),
\begin{alignat*}{3}
z^{\mathrm{h}(-\{i,j\})}_{st} &= \frac{z_{st} - z^{\mathrm{h}(-j)}_{si} z_{it}  - z^{\mathrm{h}(-i)}_{sj} z_{jt} }{z_{tt} - z^{\mathrm{h}(-j)}_{ti} z_{it}  - z^{\mathrm{h}(-i)}_{tj} z_{jt} } &&\quad \text{(obtained from (R.8.1))} \\
&= \frac{z^{(-i)}_{st} - z^{\mathrm{h}(-i)}_{sj} z^{(-i)}_{jt}}{z^{(-i)}_{tt} - z^{\mathrm{h}(-i)}_{tj} z^{(-i)}_{jt}}, &&\quad \text{(obtained from (R.8.2))} \\
&= \frac{z^{(-j)}_{st} - z^{\mathrm{h}(-j)}_{si} z^{(-j)}_{it}}{z^{(-j)}_{tt} - z^{\mathrm{h}(-j)}_{ti} z^{(-j)}_{it}} &&\quad \text{(obtained from (R.8.3))}
\end{alignat*}
and by dividing numerators and denominators by, respectively, $z_{tt}$, $z^{(-i)}_{tt}$ and $z^{(-j)}_{tt}$, we get the results (R.9.1)-(R.9.3).

\paragraph{(R.10).} This result can be obtained by developing expressions in terms of $z_{kl}$ and $z^\mathrm{h}_{kl}$, and grouping them differently. However, we will use a more intuitive argument here, similar to (R.7). Assuming $t \neq i \neq j \neq t$, let $\mathcal{P}^{\mathrm{h}(i<j)}_{st}$ be the set of all hitting paths containing $i$ and $j$ with node $i$ appearing first (before node $j$) on the paths. We have
\begin{align*}
w\left(\mathcal{P}^{\mathrm{h}(+\{i,j\})}_{st}\right) = w\left(\mathcal{P}^{\mathrm{h}(i<j)}_{st}\right) + w\left(\mathcal{P}^{\mathrm{h}(j<i)}_{st}\right)
\end{align*}
As for (R.7), observe that there is a bijection between $\mathcal{P}^{\mathrm{h}(i<j)}_{st}$ and $(\mathcal{P}^{\mathrm{h}(-\{j,t\})}_{si} \circ \mathcal{P}^{\mathrm{h}(-t)}_{ij} \circ \mathcal{P}^{\mathrm{h}}_{jt})$, with corresponding weights. Thus, 
\begin{align*}
w\left(\mathcal{P}^{\mathrm{h}(i<j)}_{st}\right) = w\left(\mathcal{P}^{\mathrm{h}(-\{j,t\})}_{si}\right) w\left(\mathcal{P}^{\mathrm{h}(-t)}_{ij}\right) w\left(\mathcal{P}^{\mathrm{h}}_{jt}\right) = z^{\mathrm{h}(-\{j,t\})}_{si} z^{\mathrm{h}(-t)}_{ij} z^{\mathrm{h}}_{jt}.
\end{align*}
Moreover, from (R.6),
\begin{align*}
w\left(\mathcal{P}^{\mathrm{h}(+\{i,j\})}_{st}\right) &=  z^{\mathrm{h}(-\{j,t\})}_{si} z^{\mathrm{h}(-t)}_{ij} z^{\mathrm{h}}_{jt} + z^{\mathrm{h}(-\{i,t\})}_{sj} z^{\mathrm{h}(-t)}_{ji} z^{\mathrm{h}}_{it} \\
&= z^{\mathrm{h}(-\{j,t\})}_{si}  z^{\mathrm{h}(+j)}_{it} + z^{\mathrm{h}(-\{i,t\})}_{sj} z^{\mathrm{h}(+i)}_{jt}. \tag{R.10}
\end{align*}
When $j = t$, $w\big(\mathcal{P}^{\mathrm{h}(+\{i,j\})}_{st}\big) = w\big(\mathcal{P}^{\mathrm{h}(+i)}_{st}\big) = z^{\mathrm{h}(-t)}_{si} z^\mathrm{h}_{it}$, from (R.6). It corresponds to the formula above knowing that the first term gives $z^{\mathrm{h}(-\{t,t\})}_{si} z^{\mathrm{h}(+t)}_{it} = z^{\mathrm{h}(-t)}_{si} z^{\mathrm{h}}_{it}$ and that the second term is zero. The same applies for $i = t$.

\subsection{Results involving any number of intermediary nodes}

\paragraph{(R.11).} Recall that $\mathcal{S}_i = \mathcal{I} \setminus i, \, \forall i \in \mathcal{I}$ or, in other words, $\mathcal{I} = \mathcal{S}_i \cup \{ i \}$. Similarly to the proof of (R.3), we notice that there is a bijection between the set of paths $\mathcal{P}^{\mathrm{h}(-\mathcal{S}_i)}_{si} \circ \mathcal{P}^{(-\mathcal{S}_i)}_{it}$ and the set of paths in $\mathcal{P}^{(-\mathcal{S}_i)}_{st}$ \emph{containing} $i$. Thus,
\begin{align*}
w\left(\mathcal{P}^{(-\mathcal{I})}_{st}\right) &= w\left(\mathcal{P}^{(-\mathcal{S}_i \cup \{ i \})}_{st}\right) = \sum_{\wp^{(-\mathcal{S}_i)}_{st} \in \mathcal{P}^{(-\mathcal{S}_i)}_{st}} \big( 1 - \delta(i \in \wp^{(-\mathcal{S}_i)}_{st}) \big) \, w\big(\wp^{(-\mathcal{S}_i)}_{st}\big) \\
&= z^{(-\mathcal{S}_i)}_{st} -  \sum_{\wp^{\mathrm{h}(-\mathcal{S}_i)}_{si} \in \mathcal{P}^{\mathrm{h}(-\mathcal{S}_i)}_{si}} \sum_{\wp^{(-\mathcal{S}_i)}_{it} \in \mathcal{P}^{(-\mathcal{S}_i)}_{it}}  w\big(\wp^{\mathrm{h}(-\mathcal{S}_i)}_{si}\big) w\big(\wp^{(-\mathcal{S}_i)}_{it}\big) \\
&= z^{(-\mathcal{S}_i)}_{st} - z^{\mathrm{h}(-\mathcal{S}_i)}_{si} z^{(-\mathcal{S}_i)}_{it}. \tag{R.11}
\end{align*}

\paragraph{(R.12).} Similarly to (R.2), with the bijection between $(\mathcal{P}^{\mathrm{h}(-\mathcal{I})}_{st} \circ \mathcal{P}^{(-\mathcal{I})}_{tt})$ and $\mathcal{P}^{(-\mathcal{I})}_{st}$, we have
\begin{align*}
w\left(\mathcal{P}^{(-\mathcal{I})}_{st}\right) = w\left(\mathcal{P}^{\mathrm{h}(-\mathcal{I})}_{st}\right) w\left(\mathcal{P}^{(-\mathcal{I})}_{tt}\right) = z^{\mathrm{h}(-\mathcal{I})}_{st} z^{(-\mathcal{I})}_{tt}.
\end{align*}
So, $z^{\mathrm{h}(-\mathcal{I})}_{st} =  z^{(-\mathcal{I})}_{st}/z^{(-\mathcal{I})}_{tt}$, and we get the result (R.12) by using (R.11) and dividing the numerator and the denominator by $z^{(-\mathcal{S}_i)}_{tt}$.

\paragraph{(R.13) -- (R.14).} We already know how to calculate weights of sets of paths avoiding a set of nodes. Conversely, let us now compute weights on path containing a set of nodes $\mathcal{I} = \mathcal{S}_i \cup \{ i \}$ from these quantities. Indeed, $z^{(+\mathcal{I})}_{st} = w\big(\mathcal{P}^{(+\mathcal{I})}_{st}\big) = w\big(\mathcal{P}^{(+\mathcal{S}_i \cup \{ i \})}_{st}\big)$ can be obtained through the \emph{inclusion-exclusion principle} (see, e.g., \cite{Brualdi-2009}) as follows. In this context, the properties that are studied are the (mutually exclusive) presence or absence of nodes on the paths.

Observe that the total weight of the union of all sets of paths from $s$ to $t$ \emph{avoiding at least one node in} $\mathcal{I}$, i.e., $w\big( \bigcup_{i \in \mathcal{I}} \mathcal{P}^{(-i)}_{st} \big)$, can be computed by (see \cite{Brualdi-2009}, Eq. (6.3))
\begin{align*}
w\left( {\textstyle \bigcup_{i \in \mathcal{I}}} \mathcal{P}^{(-i)}_{st} \right) &= \sum_{\mathcal{S} \subseteq \mathcal{I} \atop \mathcal{S} \ne \varnothing} (-1)^{(|\mathcal{S}|-1)} w \left( \mathcal{P}^{(-S)}_{st} \right) = - \sum_{\mathcal{S} \subseteq \mathcal{I} \atop \mathcal{S} \ne \varnothing} (-1)^{|\mathcal{S}|} w \left( \mathcal{P}^{(-S)}_{st} \right),
\end{align*}
where the summation on $\mathcal{S} \subseteq \mathcal{I}$ with $\mathcal{S} \ne \varnothing$ means a summation on all subsets $\mathcal{S}$ of $\mathcal{I}$, except the empty set.
In the last equation, if we denote some subset $\mathcal{S} = \{ i_{1}, i_{2}, \dots, i_{m} \}$ (where all nodes are distinct) then the set of paths avoiding all nodes in $\mathcal{S}$ is $\mathcal{P}^{(-S)}_{st} = \mathcal{P}^{(-i_{1})}_{st} \cap \mathcal{P}^{(-i_{2})}_{st} \cap \ldots \cap \mathcal{P}^{(-i_{m})}_{st} = \bigcap_{i \in \mathcal{S}} \mathcal{P}^{(-i)}_{st}$.

Now it is clear that the set of paths visiting all nodes in $\mathcal{I}$ (the set we are interested in) is equal to the set of all paths between $s$ and $t$ minus the set of paths avoiding at least one node in $\mathcal{I}$ (the quantity in the last equation). In other words, $w\big(\mathcal{P}_{st}\big) = w\big(\mathcal{P}^{(+\mathcal{I})}_{st}\big) + w\big( \bigcup_{i \in \mathcal{I}} \mathcal{P}^{(-i)}_{st} \big)$.

Therefore, from the additivity of weights on disjoint subsets, $w\big(\mathcal{P}_{st}\big) = w\big(\mathcal{P}^{(+\mathcal{I})}_{st}\big) + w\big( \bigcup_{i \in \mathcal{I}} \mathcal{P}^{(-i)}_{st} \big)$. We deduce that $w\big(\mathcal{P}^{(+\mathcal{I})}_{st}\big)  = w\big(\mathcal{P}_{st}\big) - w\big( \bigcup_{i \in \mathcal{I}} \mathcal{P}^{(-i)}_{st} \big)$; thus, knowing that $w \big( \mathcal{P}_{st} \big) = z_{st}$ and $w \big( \mathcal{P}^{(-S)}_{st} \big) = z^{(-S)}_{st}$, we get the result (R.13).
The same reasoning applies to hitting paths in order to derive (R.14).

\section{Proofs of results of Section \ref{section_pass_formulae}}
\label{App_B}

First, let us recall that the number of \emph{occurrences of an edge} $(i,j)$ on path $\wp$ is $\eta((i,j) \in \wp_{st})$. Observe that \emph{node occurrences} can be computed from edge occurrences by using either
\begin{equation}
\eta(i \in \wp_{st}) = \sum_{j \in \mathcal{V}} \eta((i, j) \in \wp_{st}) + \delta_{it} = \sum_{j \in \mathcal{V}} \eta((j,i) \in \wp_{st}) + \delta_{is},
\label{nij_to_ni}
\end{equation}
where $\delta_{it}$ is the Kronecker delta. As a matter of fact, the total number of visits will be missing one unit if the node $i$ is at the end (resp. the beginning) of the path as only outgoing (resp. ingoing) edges are counted in (\ref{nij_to_ni}). Notice that these expressions are also valid for hitting paths.

Thus, in order to compute $\eta(i \in \wp_{st})$, we need to compute $\eta((i,j) \in \wp_{st})$ in function of the elements of the fundamental matrix $\mathbf{Z}$. To this end, let us first prove a preliminary result.

\subsection{Number of occurrences in terms of partial derivatives of path weights}

From the definition of $w(\wp_{st})$ (Equation (\ref{Eq_def_path_weight01})),
\begin{equation}
w(\wp_{st}) = \prod_{\tau = 0}^{\ell - 1} w_{i_{\tau} i_{\tau+1}} = \prod_{\alpha, \beta = 1}^{n} (w_{\alpha \beta})^{\eta((\alpha,\beta) \in \wp_{st})}. \nonumber
\end{equation}
Taking the partial derivative of this expression provides
\begin{align}
\frac{\partial(w(\wp_{st}))}{\partial w_{ij}} &= \eta((i,j) \in \wp_{st}) \, (w_{i j})^{\eta((i,j) \in \wp_{st}) - 1} \prod_{\substack{\alpha,\beta = 1\\ (\alpha, \beta) \neq (i,j)}}^{n} (w_{\alpha \beta})^{\eta((\alpha,\beta) \in \wp_{st})} \nonumber \\
&= \eta((i,j) \in \wp_{st}) \, \frac{ (w_{i j})^{\eta((i,j) \in \wp_{st})} }{ w_{i j} } \prod_{\substack{\alpha,\beta = 1\\ (\alpha, \beta) \neq (i,j)}}^{n} (w_{\alpha \beta})^{\eta((\alpha,\beta) \in \wp_{st})} \nonumber \\
&= \eta((i,j) \in \wp_{st}) \frac{w(\wp_{st})}{w_{ij}}.
\label{Eq_first_derivative01} 
\end{align}



From this last result, the following relationship holds
\begin{equation}
\eta((i,j) \in \wp_{st}) \, w(\wp_{st}) = w_{ij} \frac{\partial(w(\wp_{st}))}{\partial w_{ij}}
\label{nij_to_pder}.
\end{equation}

Taking once more the partial derivative of $w(\wp_{st})$ with respect to $w_{kl}$, $(k,l) \ne (i,j)$, yields
\begin{align}
\frac{\partial^{2}(w(\wp_{st}))}{\partial w_{kl} \partial w_{ij}}
&= \eta((i,j) \in \wp_{st}) \, (w_{i j})^{\eta((i,j) \in \wp_{st}) - 1} \eta((k,l) \in \wp_{st}) \, (w_{k l})^{\eta((k,l) \in \wp_{st}) - 1} \nonumber \\
&\quad \times \prod_{\substack{\alpha,\beta = 1\\ (\alpha, \beta) \neq \{ (i,j),(k,l) \} }}^{n} (w_{\alpha \beta})^{\eta((\alpha,\beta) \in \wp_{st})} \nonumber \\
&= \eta((i,j) \in \wp_{st})  \eta((k,l) \in \wp_{st}) \frac{w(\wp_{st})}{w_{ij} w_{kl}}.
\end{align}

Therefore, for $(k,l) \ne (i,j)$,
\begin{align}
\eta((i,j) \in \wp_{st}) \eta((k,l) \in \wp_{st}) \, w(\wp_{st}) &= w_{ij} w_{kl} \frac{\partial^{2}(w(\wp_{st}))}{\partial w_{kl} \partial w_{ij}}. \nonumber
\end{align}

Now, proceeding in the same way for $(i,j) = (k,l)$, we obtain from Equation (\ref{Eq_first_derivative01})
\begin{equation}
\frac{\partial^{2}(w(\wp_{st}))}{(\partial w_{ij})^2}
= \eta((i,j) \in \wp_{st})  \eta((i,j) \in \wp_{st}) \frac{w(\wp_{st})}{(w_{ij})^2}
- \eta((i,j) \in \wp_{st}) \frac{w(\wp_{st})}{(w_{ij})^2}
\end{equation}

After multiplying this last equation by $(w_{ij})^{2}$ and then using (\ref{nij_to_pder}), this provides an additional term, leading to the following, more general, expression,
\begin{equation}
\eta((i,j) \in \wp_{st}) \eta((k,l) \in \wp_{st}) \, w(\wp_{st})
= w_{ij} w_{kl} \frac{\partial^{2}(w(\wp_{st}))}{\partial w_{kl} \partial w_{ij}}
\quad + \delta_{ik}  \delta_{jl} w_{ij} \frac{\partial(w(\wp_{st}))}{\partial w_{ij}}.
\label{nij_to_pder2}
\end{equation}
which is also valid when $(i,j) = (k,l)$.
Equations (\ref{nij_to_pder}) and (\ref{nij_to_pder2}) are the first needed preliminary results. Note that these two expressions also hold for hitting paths.

\subsection{Number of occurrences in terms of the fundamental matrix}

From here, results for the non-hitting and hitting paths will differ. Before proceeding, we still need another useful result showing that 
\begin{align}
&\frac{\partial z_{st}}{\partial w_{ij}} = z_{si} z_{jt} \label{Eq_first_derivative_fundamental_matrix01}, \\
&\frac{\partial^{2} z_{st}}{\partial w_{kl} \partial w_{ij}} =  z_{sk} z_{li} z_{jt} +  z_{si} z_{jk} z_{lt} \label{Eq_second_derivative_fundamental_matrix01}.
\end{align}
These expressions are obtained by using the standard formula computing the partial derivative of a matrix inverse, $\frac{\partial \mathbf{M}^{-1}}{\partial m_{ij}} = - \mathbf{M}^{-1} \frac{\partial \mathbf{M}}{\partial m_{ij}} \mathbf{M}^{-1}$ (details and similar approaches can be found in \cite{Francoisse-2017,kivimaki2016two,Kivimaki-2012,Mantrach-2009}).
When applying this formula to the fundamental matrix $\mathbf{Z} = (\mathbf{I} - \mathbf{W})^{-1}$,
\begin{align}
\frac{\partial z_{st}}{\partial w_{ij}}
&= \mathbf{e}_{s}^\top \frac{\partial \mathbf{Z}}{\partial w_{ij}} \mathbf{e}_{t}
= - \mathbf{e}_{s}^\top \mathbf{Z} \frac{\partial (\mathbf{I} - \mathbf{W})}{\partial w_{ij}} \mathbf{Z} \mathbf{e}_{t} \nonumber \\
&= \mathbf{e}_{s}^\top \mathbf{Z} \mathbf{e}_{i} \mathbf{e}_{j}^\top \mathbf{Z} \mathbf{e}_{t}
= z_{si} z_{jt}, \nonumber
\end{align}
which is the first result (\ref{Eq_first_derivative_fundamental_matrix01}).
Taking once more the partial derivative provides (\ref{Eq_second_derivative_fundamental_matrix01}),
\begin{align}
\frac{\partial^{2} z_{st}}{\partial w_{kl} \partial w_{ij}}
&= \frac{\partial (z_{si} z_{jt})}{\partial w_{kl}}
= \frac{\partial (z_{si})}{\partial w_{kl}} z_{jt} + z_{si} \frac{\partial (z_{jt})}{\partial w_{kl}}
= z_{sk} z_{li} z_{jt} +  z_{si} z_{jk} z_{lt}. \nonumber
\end{align}

\noindent We are now ready to prove the next results (R.15)-(R.18).

\subsubsection{Results for regular paths}

\paragraph{(R.15) -- (R.16).}
For regular paths, following (\ref{nij_to_pder}), (R.1), and (\ref{Eq_first_derivative_fundamental_matrix01}), we find for edge occurrences
\begin{align}
\sum_{\wp_{st} \in \mathcal{P}_{st}} \eta((i,j) \in \wp_{st}) \, w(\wp_{st}) &= \sum_{\wp_{st} \in \mathcal{P}_{st}} \frac{\partial w(\wp_{st})}{\partial w_{ij}} w_{ij} = \frac{\partial \left( \sum_{\wp_{st} \in \mathcal{P}_{st}} w(\wp_{st}) \right)}{\partial w_{ij}} w_{ij} \nonumber \\
&= \frac{\partial z_{st}}{\partial w_{ij}} w_{ij} = z_{si} w_{ij} z_{jt}.
\label{Eq_first_derivative_result01}
\end{align}
Moreover, from (\ref{nij_to_pder2}), (R.1) and (\ref{Eq_first_derivative_fundamental_matrix01})-(\ref{Eq_second_derivative_fundamental_matrix01}),
\begin{align}
\sum_{\wp_{st} \in \mathcal{P}_{st}} & \eta((i,j) \in \wp_{st}) \eta((k,l) \in \wp_{st}) \, w(\wp_{st}) \notag \\
&= \sum_{\wp_{st} \in \mathcal{P}_{st}} \frac{\partial^{2} w(\wp_{st})}{\partial w_{kl} \partial w_{ij}} w_{ij} w_{kl} + \delta_{ik}  \delta_{jl}  \sum_{\wp_{st} \in \mathcal{P}_{st}} \frac{\partial w(\wp_{st})}{\partial w_{ij}} w_{ij}  \notag \\
&= \frac{\partial^{2} z_{st}}{\partial w_{kl} \partial w_{ij}} w_{ij} w_{kl} + \delta_{ik}  \delta_{jl} \frac{\partial z_{st}}{\partial w_{ij}} w_{ij} \notag \\
&=  z_{si} w_{ij} z_{jk} w_{kl} z_{lt} + z_{sk} w_{kl} z_{li} w_{ij} z_{jt} + \delta_{ik}  \delta_{jl} z_{si} w_{ij} z_{jt}.
\label{Eq_second_derivative_result01}
\end{align}

This is almost the expected result, as we are mainly interested in the closely related quantities involving node occurrences instead of edge occurrences, $\sum_{\wp_{st} \in \mathcal{P}_{st}} \eta(i \in \wp_{st}) \, w(\wp_{st})$ and $\sum_{\wp_{st} \in \mathcal{P}_{st}}  \eta(i \in \wp_{st}) \eta(k \in \wp_{st}) \, w(\wp_{st})$.
For the first expression, using (R.1), (\ref{nij_to_ni}), (\ref{Eq_first_derivative_result01}) and $(\mathbf{I} - \mathbf{W})\mathbf{Z} = \mathbf{I}$, i.e., $\sum_{i \in \mathcal{V}} w_{si} z_{it} = z_{st} - \delta_{st}$, finally provides (R.15)
\begin{align}
\sum_{\wp_{st} \in \mathcal{P}_{st}} \eta(i \in \wp_{st}) \, w(\wp_{st})
&= \sum_{\wp_{st} \in \mathcal{P}_{st}} \bigg ( \sum_{j \in \mathcal{V}} \eta((i, j) \in \wp_{st}) + \delta_{it} \bigg ) \, w(\wp_{st}) \nonumber \\
&= \sum_{j \in \mathcal{V}} \underbracket[0.5pt][5pt]{ \sum_{\wp_{st} \in \mathcal{P}_{st}} \eta((i, j) \in \wp_{st}) \, w(\wp_{st}) }_{z_{si} w_{ij} z_{jt}} + \delta_{it} \underbracket[0.5pt][5pt]{ \sum_{\wp_{st} \in \mathcal{P}_{st}} \, w(\wp_{st}) }_{z_{st}} \nonumber \\
&= \sum_{j \in \mathcal{V}} z_{si} w_{ij} z_{jt} + \delta_{it} z_{st} \nonumber \\
&= z_{si} (z_{it} - \delta_{it}) + \delta_{it} z_{st} = z_{si} z_{it}. \tag{R.15}
\end{align}

Let us now compute the second expression,
\begin{align}
&\sum_{\wp_{st} \in \mathcal{P}_{st}}  \eta(i \in \wp_{st}) \eta(k \in \wp_{st}) \, w(\wp_{st}) \nonumber \\
&= \sum_{\wp_{st} \in \mathcal{P}_{st}} \bigg ( \sum_{j \in \mathcal{V}} \eta((i, j) \in \wp_{st}) + \delta_{it} \bigg ) \bigg ( \sum_{l \in \mathcal{V}} \eta((k, l) \in \wp_{st}) + \delta_{kt} \bigg ) \, w(\wp_{st}) \nonumber \\
&= \sum_{\wp_{st} \in \mathcal{P}_{st}} \bigg ( \sum_{j \in \mathcal{V}} \eta((i, j) \in \wp_{st}) \bigg ) \bigg ( \sum_{l \in \mathcal{V}} \eta((k, l) \in \wp_{st}) \bigg ) \, w(\wp_{st}) \nonumber \\
&\quad + \sum_{\wp_{st} \in \mathcal{P}_{st}} \bigg ( \sum_{j \in \mathcal{V}} \eta((i, j) \in \wp_{st}) \bigg )  \delta_{kt} \, w(\wp_{st}) \nonumber \\
&\quad + \sum_{\wp_{st} \in \mathcal{P}_{st}} \bigg ( \sum_{l \in \mathcal{V}} \eta((k, l) \in \wp_{st}) \bigg ) \delta_{it} \, w(\wp_{st}) \nonumber \\
&\quad + \sum_{\wp_{st} \in \mathcal{P}_{st}} \delta_{it} \delta_{kt} \, w(\wp_{st}) \nonumber \\
&= \sum_{j \in \mathcal{V}} \sum_{l \in \mathcal{V}} \bigg ( \sum_{\wp_{st} \in \mathcal{P}_{st}} \eta((i, j) \in \wp_{st}) \eta((k, l) \in \wp_{st}) \, w(\wp_{st}) \bigg ) \nonumber \\
&\quad + \delta_{kt} \sum_{j \in \mathcal{V}} \bigg ( \sum_{\wp_{st} \in \mathcal{P}_{st}} \eta((i, j) \in \wp_{st}) \, w(\wp_{st}) \bigg ) \nonumber \\
&\quad + \delta_{it} \sum_{l \in \mathcal{V}} \bigg ( \sum_{\wp_{st} \in \mathcal{P}_{st}} \eta((k, l) \in \wp_{st}) \, w(\wp_{st}) \bigg ) \nonumber \\
&\quad + \delta_{it} \delta_{kt} \bigg ( \sum_{\wp_{st} \in \mathcal{P}_{st}} w(\wp_{st}) \bigg ).
\label{Eq_co-occurrences_development01}
\end{align}

From this last result, further using Equations (\ref{Eq_first_derivative_result01})-(\ref{Eq_second_derivative_result01}), (R.1) and, again, $\sum_{i \in \mathcal{V}} w_{si} z_{it} = z_{st} - \delta_{st}$ provides (R.16):
\begin{align}
\sum_{\wp_{st} \in \mathcal{P}_{st}}  &\eta(i \in \wp_{st}) \eta(k \in \wp_{st}) \, w(\wp_{st}) \nonumber \\
&= \sum_{j \in \mathcal{V}} \sum_{l \in \mathcal{V}} \left ( z_{si} w_{ij} z_{jk} w_{kl} z_{lt} + z_{sk} w_{kl} z_{li} w_{ij} z_{jt} + \delta_{ik}  \delta_{jl} z_{si} w_{ij} z_{jt} \right ) \nonumber \\
&\quad + \delta_{kt} \sum_{j \in \mathcal{V}} \left ( z_{si} w_{ij} z_{jt} \right )
+ \delta_{it} \sum_{l \in \mathcal{V}} \left ( z_{sk} w_{kl} z_{lt} \right )
+ \delta_{it} \delta_{kt} \, z_{st} \nonumber \\
&= \big ( z_{si} (z_{ik} - \delta_{ik}) (z_{kt} - \delta_{kt}) + z_{sk} (z_{ki} - \delta_{ki}) (z_{it} - \delta_{it}) + \delta_{ik}  z_{si} (z_{it} - \delta_{it}) \big ) \nonumber \\
&\quad + \delta_{kt} z_{si} (z_{it} - \delta_{it})
+ \delta_{it} z_{sk} (z_{kt} - \delta_{kt})
+ \delta_{it} \delta_{kt} \, z_{st} \nonumber \\
&= z_{si}z_{ik}z_{kt} + z_{sk}z_{ki}z_{it} - \delta_{ik} z_{si} z_{kt}. \tag{R.16}
\end{align}

\subsubsection{Results for hitting paths}

\paragraph{(R.17) -- (R.18).} We proceed in the same way for hitting paths. By using the expressions (\ref{Eq_first_derivative_fundamental_matrix01})-(\ref{Eq_second_derivative_fundamental_matrix01}) for computing the partial derivative of $z^\mathrm{h}_{st} = z_{st}/z_{tt}$ (see Equation (R.2)), and using (R.4), we get
\begin{equation}
\frac{\partial z^\mathrm{h}_{st}}{\partial w_{ij}}
= \frac{\partial z_{st}}{\partial w_{ij}} z_{tt}^{-1} + z_{st} \frac{\partial z_{tt}^{-1}}{\partial w_{ij}}
= \left( z_{si} - z^\mathrm{h}_{st} z_{ti} \right)z^\mathrm{h}_{jt} = z^{(-t)}_{si} z^\mathrm{h}_{jt},
\end{equation}
\begin{align}
 \frac{\partial^{2} z^\mathrm{h}_{st}}{\partial w_{kl} \partial w_{ij}} &= (z_{si} - z^\mathrm{h}_{st} z_{ti}) (z_{jk} - z^\mathrm{h}_{jt} z_{tk}) z^\mathrm{h}_{lt} + (z_{sk} - z^\mathrm{h}_{st} z_{tk}) (z_{li} - z^\mathrm{h}_{lt} z_{ti}) z^\mathrm{h}_{jt} \notag \\
 &= z^{(-t)}_{si} z^{(-t)}_{jk} z^\mathrm{h}_{lt} + z^{(-t)}_{sk} z^{(-t)}_{li} z^\mathrm{h}_{jt}.
\end{align}
As for the previous subsection (see Equations (\ref{Eq_first_derivative_result01})-(\ref{Eq_second_derivative_result01})), by using (\ref{nij_to_pder})-(\ref{nij_to_pder2}), (R.2), and the previous expressions for the partial derivatives of $z^\mathrm{h}_{st}$, we obtain the equivalent of (\ref{Eq_first_derivative_result01}) and (\ref{Eq_second_derivative_result01}) for hitting paths,
\begin{equation}
\sum_{\wp^\mathrm{h}_{st} \in \mathcal{P}^\mathrm{h}_{st}} \eta((i,j) \in \wp^\mathrm{h}_{st}) \, w(\wp^\mathrm{h}_{st}) = \sum_{\wp^\mathrm{h}_{st} \in \mathcal{P}^\mathrm{h}_{st}} \frac{\partial w(\wp^\mathrm{h}_{st})}{\partial w_{ij}} w_{ij}
= \frac{\partial z^\mathrm{h}_{st}}{\partial w_{ij}} w_{ij}
= z^{(-t)}_{si} w_{ij} z^\mathrm{h}_{jt},
\label{Eq_first_derivative_hitting_result01}
\end{equation}
and
\begin{align}
\sum_{\wp^\mathrm{h}_{st} \in \mathcal{P}^\mathrm{h}_{st}} & \eta((i,j) \in \wp^\mathrm{h}_{st}) \eta((k,l) \in \wp^\mathrm{h}_{st}) \, w(\wp^\mathrm{h}_{st}) \notag \\
&= \sum_{\wp^\mathrm{h}_{st} \in \mathcal{P}^\mathrm{h}_{st}} \frac{\partial w(\wp^\mathrm{h}_{st})}{\partial w_{kl} \partial w_{ij}} w_{ij} w_{kl} + \delta_{ik}  \delta_{jl}  \sum_{\wp^\mathrm{h}_{st} \in \mathcal{P}^\mathrm{h}_{st}} \frac{\partial w(\wp^\mathrm{h}_{st})}{\partial w_{ij}} w_{ij},  \notag \\
&= \frac{\partial z^\mathrm{h}_{st}}{\partial w_{kl} \partial w_{ij}} w_{ij} w_{kl} + \delta_{ik}  \delta_{jl} \frac{\partial z^\mathrm{h}_{st}}{\partial w_{ij}} w_{ij} \notag \\
&=  z^{(-t)}_{si} w_{ij} z^{(-t)}_{jk} w_{kl} z^\mathrm{h}_{lt} + z^{(-t)}_{sk} w_{kl} z^{(-t)}_{li} w_{ij} z^\mathrm{h}_{jt} + \delta_{ik}  \delta_{jl} z^{(-t)}_{si} w_{ij} z^\mathrm{h}_{jt},
\label{Eq_second_derivative_hitting_result01}
\end{align}
These results are surprisingly similar to the equivalent results for regular paths, displayed in Equations (\ref{Eq_first_derivative_result01}) and (\ref{Eq_second_derivative_result01}).

But, as before, we are in fact interested in the closely related quantities involving node occurrences, $\sum_{\wp_{st}^{\mathrm{h}} \in \mathcal{P}^{\mathrm{h}}_{st}} \eta(i \in \wp^{\mathrm{h}}_{st}) \, w(\wp^{\mathrm{h}}_{st})$ and $\sum_{\wp^{\mathrm{h}}_{st} \in \mathcal{P}^{\mathrm{h}}_{st}}  \eta(i \in \wp^{\mathrm{h}}_{st}) \eta(k \in \wp^{\mathrm{h}}_{st}) \, w(\wp^{\mathrm{h}}_{st})$.
For the first expression, using (R.1), (R.2), (\ref{nij_to_ni}), (\ref{Eq_first_derivative_hitting_result01}), and $\sum_{i \in \mathcal{V}} w_{si} z_{it} = z_{st} - \delta_{st}$, finally provides result (R.17):
\begin{align}
\sum_{\wp^\mathrm{h}_{st} \in \mathcal{P}^\mathrm{h}_{st}} \eta(i \in \wp^\mathrm{h}_{st}) \, w(\wp^\mathrm{h}_{st})
&= \sum_{\wp^\mathrm{h}_{st} \in \mathcal{P}^\mathrm{h}_{st}} \bigg ( \sum_{j \in \mathcal{V}} \eta((i, j) \in \wp^\mathrm{h}_{st}) + \delta_{it} \bigg ) \, w(\wp^\mathrm{h}_{st}) \nonumber \\
&= \sum_{j \in \mathcal{V}} \underbracket[0.5pt][5pt]{ \sum_{\wp^\mathrm{h}_{st} \in \mathcal{P}^\mathrm{h}_{st}} \eta((i, j) \in \wp^\mathrm{h}_{st}) \, w(\wp^\mathrm{h}_{st}) }_{z^{(-t)}_{si} w_{ij} z^\mathrm{h}_{jt}} + \delta_{it} \underbracket[0.5pt][5pt]{ \sum_{\wp^\mathrm{h}_{st} \in \mathcal{P}^\mathrm{h}_{st}} \, w(\wp^\mathrm{h}_{st}) }_{z^\mathrm{h}_{st}} \nonumber \\
&= \sum_{j \in \mathcal{V}} z^{(-t)}_{si} w_{ij} z^\mathrm{h}_{jt} + \delta_{it} z^\mathrm{h}_{st}
= z^{(-t)}_{si} \sum_{j \in \mathcal{V}} w_{ij} \frac{z_{jt}}{z_{tt}} + \delta_{it} z^\mathrm{h}_{st} \nonumber \\
&= z^{(-t)}_{si} \frac{ (z_{it} - \delta_{it}) }{z_{tt}} + \delta_{it} z^\mathrm{h}_{st}
= z^{(-t)}_{si} z^\mathrm{h}_{it} + \delta_{it} z^\mathrm{h}_{st}, \tag{R.17}
\end{align}
where we used $z^{(-t)}_{si} \delta_{it} = 0$ (see (R.5)) in the last equality.

Let us now compute the second expression by proceeding in the same way as in the previous section for non-hitting paths (see Equation (\ref{Eq_co-occurrences_development01})),
\begin{align}
&\sum_{\wp^\mathrm{h}_{st} \in \mathcal{P}^\mathrm{h}_{st}}  \eta(i \in \wp^\mathrm{h}_{st}) \eta(k \in \wp^\mathrm{h}_{st}) \, w(\wp^\mathrm{h}_{st}) \nonumber \\
&\qquad = \sum_{\wp^\mathrm{h}_{st} \in \mathcal{P}^\mathrm{h}_{st}} \bigg ( \sum_{j \in \mathcal{V}} \eta((i, j) \in \wp^\mathrm{h}_{st}) + \delta_{it} \bigg ) \bigg ( \sum_{l \in \mathcal{V}} \eta((k, l) \in \wp^\mathrm{h}_{st}) + \delta_{kt} \bigg ) \, w(\wp^\mathrm{h}_{st}) \nonumber \\
&\qquad = \sum_{\wp^\mathrm{h}_{st} \in \mathcal{P}^\mathrm{h}_{st}} \bigg ( \sum_{j \in \mathcal{V}} \eta((i, j) \in \wp^\mathrm{h}_{st}) \bigg ) \bigg ( \sum_{l \in \mathcal{V}} \eta((k, l) \in \wp^\mathrm{h}_{st}) \bigg ) \, w(\wp^\mathrm{h}_{st}) \nonumber \\
&\qquad \quad + \sum_{\wp^\mathrm{h}_{st} \in \mathcal{P}^\mathrm{h}_{st}} \bigg ( \sum_{j \in \mathcal{V}} \eta((i, j) \in \wp^\mathrm{h}_{st}) \bigg ) \delta_{kt} \, w(\wp^\mathrm{h}_{st}) \nonumber \\
&\qquad \quad + \sum_{\wp^\mathrm{h}_{st} \in \mathcal{P}^\mathrm{h}_{st}} \bigg ( \sum_{l \in \mathcal{V}} \eta((k, l) \in \wp^\mathrm{h}_{st}) \bigg ) \delta_{it} \, w(\wp^\mathrm{h}_{st}) \nonumber \\
&\qquad \quad + \sum_{\wp^\mathrm{h}_{st} \in \mathcal{P}^\mathrm{h}_{st}} \delta_{it} \delta_{kt} \, w(\wp^\mathrm{h}_{st}).
\end{align}
Further using Equations (\ref{Eq_first_derivative_hitting_result01})-(\ref{Eq_second_derivative_hitting_result01}), (R.1), (R.4) provides
\begin{align}
&\sum_{\wp^\mathrm{h}_{st} \in \mathcal{P}^\mathrm{h}_{st}} \eta(i \in \wp^\mathrm{h}_{st}) \eta(k \in \wp^\mathrm{h}_{st}) \, w(\wp^\mathrm{h}_{st}) \nonumber \\
&= \sum_{j,l \in \mathcal{V}} \left ( z^{(-t)}_{si} w_{ij} z^{(-t)}_{jk} w_{kl} z^\mathrm{h}_{lt} + z^{(-t)}_{sk} w_{kl} z^{(-t)}_{li} w_{ij} z^\mathrm{h}_{jt} + \delta_{ik}  \delta_{jl} z^{(-t)}_{si} w_{ij} z^\mathrm{h}_{jt} \right ) \nonumber \\
&\quad + \delta_{kt} \sum_{j \in \mathcal{V}} \left ( z^{(-t)}_{si} w_{ij} z^\mathrm{h}_{jt} \right )
+ \delta_{it} \sum_{l \in \mathcal{V}} \left ( z^{(-t)}_{sk} w_{kl} z^\mathrm{h}_{lt} \right )
+ \delta_{it} \delta_{kt} \, z^\mathrm{h}_{st} \nonumber \\
&= z^{(-t)}_{si} \Big( \sum_{j \in \mathcal{V}}  w_{ij} z^{(-t)}_{jk} \Big( \sum_{l \in \mathcal{V}} w_{kl} z^\mathrm{h}_{lt} \Big) \Big) + z^{(-t)}_{sk} \Big( \sum_{l \in \mathcal{V}} w_{kl} z^{(-t)}_{li} \Big( \sum_{j \in \mathcal{V}} w_{ij} z^\mathrm{h}_{jt} \Big) \Big) \nonumber \\
&\quad + \delta_{ik} z^{(-t)}_{si} \Big( \sum_{j \in \mathcal{V}} w_{ij} z^\mathrm{h}_{jt} \Big)
+ \delta_{kt} z^{(-t)}_{si}  \Big( \sum_{j \in \mathcal{V}} w_{ij} z^\mathrm{h}_{jt} \Big)
+ \delta_{it} z^{(-t)}_{sk} \Big( \sum_{l \in \mathcal{V}} w_{kl} z^\mathrm{h}_{lt} \Big) \nonumber \\
&\quad + \delta_{it} \delta_{kt} \, z^\mathrm{h}_{st}.
\label{Eq_derivation_node_occurrences01}
\end{align}
Before going further, note that we encounter expressions like $\sum_{j \in \mathcal{V}} w_{ij} z^\mathrm{h}_{jt}$ and $\sum_{j \in \mathcal{V}}  w_{ij} z^{(-t)}_{jt}$, each multiplied by $z^{(-t)}_{si}$. Let us compute these expressions.
We already know that $(\mathbf{I} - \mathbf{W})\mathbf{Z} = \mathbf{I}$, i.e., $\sum_{j \in \mathcal{V}} w_{sj} z_{jt} = z_{st} - \delta_{st}$. Therefore, from (R.2),
\begin{equation}
z^{(-t)}_{si} \sum_{j \in \mathcal{V}} w_{ij} z^\mathrm{h}_{jt}
= \frac{z^{(-t)}_{si}}{z_{tt}} \sum_{j \in \mathcal{V}} w_{ij} z_{jt}
= \frac{z^{(-t)}_{si}}{z_{tt}} (z_{it} - \delta_{it})
= \frac{z^{(-t)}_{si}}{z_{tt}} z_{it} 
= z^{(-t)}_{si} z^\mathrm{h}_{it},
\label{Eq_intermediate_result_summation01}
\end{equation}
and the last equality holds because $z^{(-t)}_{si}$ is equal to $0$ when $i = t$ (see Equation (R.4) and the following discussion) so that $\delta_{it} z^{(-t)}_{si} = 0$. It means that this simplification is only valid when $z^{(-t)}_{si}$ is present.
Moreover, following Equation (R.4) and using the last expression (\ref{Eq_intermediate_result_summation01}),
\begin{align}
z^{(-t)}_{si} \sum_{j \in \mathcal{V}}  w_{ij} z^{(-t)}_{jk}
&= z^{(-t)}_{si} \sum_{j \in \mathcal{V}}  w_{ij} ( z_{jk} - z^{\mathrm{h}}_{jt} z_{tk} ) \nonumber \\
&= z^{(-t)}_{si} \Big( \sum_{j \in \mathcal{V}}  w_{ij} z_{jk} - z_{tk} \sum_{j \in \mathcal{V}} w_{ij} z^{\mathrm{h}}_{jt} \Big) \nonumber \\
&= z^{(-t)}_{si} \big( (z_{ik} - \delta_{ik}) - z_{tk} z^{\mathrm{h}}_{it} \big) \nonumber \\
&= z^{(-t)}_{si} \big( z^{(-t)}_{ik} - \delta_{ik} \big).
\label{Eq_intermediate_result_summation02}
\end{align}
Injecting successively the first expression (\ref{Eq_intermediate_result_summation01}) and then the second expression (\ref{Eq_intermediate_result_summation02}) in Equation (\ref{Eq_derivation_node_occurrences01}) provides (R.18),
\begin{align}
&\sum_{\wp^\mathrm{h}_{st} \in \mathcal{P}^\mathrm{h}_{st}} \eta(i \in \wp^\mathrm{h}_{st}) \eta(k \in \wp^\mathrm{h}_{st}) \, w(\wp^\mathrm{h}_{st}) \nonumber \\
& \qquad = z^{(-t)}_{si} z^\mathrm{h}_{kt} \sum_{j \in \mathcal{V}} \big ( w_{ij} z^{(-t)}_{jk} \big ) + z^{(-t)}_{sk} z^\mathrm{h}_{it} \sum_{l \in \mathcal{V}} \big ( w_{kl} z^{(-t)}_{li} \big ) \nonumber \\
& \qquad \quad + \delta_{ik} z^{(-t)}_{si} z^\mathrm{h}_{it} + \delta_{kt} z^{(-t)}_{si} z^\mathrm{h}_{it}
+ \delta_{it} z^{(-t)}_{sk} z^\mathrm{h}_{kt}
+ \delta_{it} \delta_{kt} \, z^\mathrm{h}_{st} \nonumber \\
& \qquad = z^{(-t)}_{si} \big( z^{(-t)}_{ik} - \delta_{ik} \big) z^\mathrm{h}_{kt} + z^{(-t)}_{sk} \big( z^{(-t)}_{ki} - \delta_{ki} \big) z^\mathrm{h}_{it} \notag \\
& \qquad \quad + \delta_{ik} z^{(-t)}_{si} z^\mathrm{h}_{kt} + \delta_{kt} z^{(-t)}_{si} z^\mathrm{h}_{it} + \delta_{it} z^{(-t)}_{sk} z^\mathrm{h}_{kt} + \delta_{it} \delta_{kt} z^\mathrm{h}_{st} \nonumber \\
& \qquad = z^{(-t)}_{si} z^{(-t)}_{ik} z^\mathrm{h}_{kt} + z^{(-t)}_{sk} z^{(-t)}_{ki} z^\mathrm{h}_{it} \notag \\
& \qquad \quad - \delta_{ik} z^{(-t)}_{si} z^\mathrm{h}_{kt} + \delta_{kt} z^{(-t)}_{si} z^\mathrm{h}_{it} + \delta_{it} z^{(-t)}_{sk} z^\mathrm{h}_{kt} + \delta_{it} \delta_{kt} z^\mathrm{h}_{st}. \tag{R.18}
\end{align}
Note that, contrary to regular paths (see Equation (R.16)), in this case, some terms containing Kronecker deltas do not cancel out, leading to a more complex expression.

\begin{center}
\rule{2.5in}{0.01in} 
\par\end{center}


\bibliographystyle{abbrv}
\bibliography{generalizedBoP}

\end{document}